\pdfoutput=1

\documentclass[11pt]{article}

\usepackage{acl}

\usepackage{times}
\usepackage{latexsym}

\usepackage[T1]{fontenc}

\usepackage[utf8]{inputenc}

\usepackage{microtype}

\usepackage{url}
\usepackage{todonotes}
\usepackage{xcolor}
\usepackage{tabularx}
\usepackage{booktabs}
\usepackage{caption}
\usepackage{subcaption}
\usepackage{tikz-dependency}
\usepackage{comment}
\usepackage{multirow}
\usepackage{makecell}
\usepackage{nicefrac}
\usepackage{xfrac}
\usepackage{amssymb}
\usepackage[flushleft]{threeparttable}
\usepackage{soul}

\newcommand{\scierc}{{\scshape SciERC}}
\newcommand{\semeval}{{\scshape SemEval}}

\newcommand*\circled[1]{\tikz[baseline=(char.base)]{\node[shape=circle,draw,inner sep=2pt] (char) {#1};}}

\title{\textit{What Do You Mean by Relation Extraction?} \\A Survey on Datasets and Study on Scientific Relation Classification}

\author{
Elisa Bassignana\textsuperscript{$\clubsuit$} \and
Barbara Plank\textsuperscript{$\clubsuit$}\textsuperscript{$\diamondsuit$} \\
\textsuperscript{$\clubsuit$}Department of Computer Science, 
        IT University of Copenhagen, Denmark \\
\textsuperscript{$\diamondsuit$}Center for Information and Language Processing (CIS), LMU Munich, Germany \\
\texttt{\{elba, bapl\}@itu.dk}}

\begin{document}
\maketitle
\begin{abstract}

Over the last five years, research on Relation Extraction (RE) witnessed extensive progress with many new dataset releases. At the same time, setup clarity has decreased, contributing to increased difficulty of reliable empirical evaluation~\citep{taille-etal-2020-lets}. In this paper, we provide a comprehensive survey of RE datasets, and revisit the task definition and its adoption by the community.  We find that cross-dataset and cross-domain setups are particularly lacking. 
We present an empirical study on scientific Relation Classification across two datasets. Despite large data overlap, our analysis reveals  substantial discrepancies in annotation. Annotation discrepancies strongly impact Relation Classification performance, explaining large drops in cross-dataset evaluations. Variation within further sub-domains exists but impacts Relation Classification only to limited degrees. Overall, our study calls for more rigour in reporting setups in RE and evaluation across multiple test sets. 
\end{abstract}

\section{Introduction}

\begin{figure}
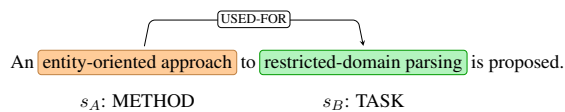

\resizebox{\columnwidth}{!}{
\begin{dependency}
\begin{deptext}
An \& entity-oriented approach \& to  \& restricted-domain parsing \& is proposed.\\ \\
 \& $s_A$: METHOD \& \& $s_B$: TASK \& \\
\end{deptext}
\wordgroup[group style={fill=orange!40, draw=brown}]{1}{2}{2}{entity1}
\wordgroup[group style={fill=green!85!blue!30!white, draw=green!60!black}]{1}{4}{4}{entity2}
\groupedge{entity1}{entity2}{USED-FOR}{3ex}
\end{dependency}
}
\caption{\label{fig:sample-re}RE annotation sample. The sentence contains two annotated spans denoting two entities, with respective types \texttt{METHOD} and \texttt{TASK}, and a semantic relation between them labeled as \texttt{USED-FOR}.}
\end{figure}

Information Extraction (IE) is a key step in Natural Language Processing (NLP) to extract  information, which is useful for question answering and knowledge base population, for example.
Relation Extraction (RE) is a specific case of IE~\cite{grishman2012information} with the focus on the identification of semantic relations between entities (see Figure~\ref{fig:sample-re}).
The aim of the most typical RE setup is the extraction of informative triples from texts.
Given a sequence of tokens $[t_0, t_1 ..., t_n]$ and two entities (spans), $s_A = [t_i, \ldots, t_j]$ and $s_B = [t_u, \ldots, t_v]$, RE triples are in the form $(s_A, s_B, r)$, where $r \in R$ and $R$ is a pre-defined set of relation labels.
Because of the directionality of the relations, $(s_B, s_A, r)$ represents a different triple.

We survey existing RE datasets---outside the biomedical domain---with an additional focus on the task definition.\footnote{We refer the reader to~\citet{bio-survey} for a survey on biomedical RE and event extraction.}
Existing RE surveys mainly focus on modeling techniques~\citep{bach2007review,pawar2017relation,aydar2021neural,liu_survey_2020}.
To the best of our knowledge, we are the first to give a comprehensive overview of available RE datasets. We also revisit RE papers from the ACL community over the last five years, to identify what part(s) of the task definition recent work focuses on. As it turns out, this is often not easy to determine, which makes fair evaluation difficult. We aim to shed light on such assumptions.\footnote{\citet{pyysalo08why} discuss similar difficulties in the biomedical domain.}

Moreover, recent work in NLP has shown that single test splits and in-distribution evaluation overestimate generalization performance, arguing for the use of multiple test sets or split evaluation~\citep{gorman-bedrick-2019-need,sogaard-etal-2021-need}.
While this direction has started to be followed by other NLP tasks~\citep{petrov2012overview,pradhan-etal-2013-towards,williams-etal-2018-broad,yu-etal-2019-sparc,zhu-etal-2020-crosswoz,liu2021crossner}, for RE \textit{cross-dataset} and \textit{cross-domain} evaluation have received little attention.
We explore this direction in the scientific domain and propose to study the possible presence of distinctive \textit{sub-domains}~\cite{lippincott-etal-2010-exploring}.
Sub-domains are differences between subsets of a domain that may be expected to behave homogeneously.
Using two scientific datasets, we study to what degree: (\emph{a}) they contain overlapping data; (\emph{b}) their annotations differ; and (\emph{c}) sub-domains impact Relation Classification (RC)---the task of classifying the relation type held between a pair of entities (details in Section~\ref{sec:the-task}).

The contributions of this paper are:
\begin{itemize}
    \item To the best of our knowledge, we are the first to provide a comprehensive survey on currently available RE datasets.
    \item We define RE considering its modularity. We analyze previous works and find unclarity in setups;
    we call for more rigour in specifying which RE sub-part(s) are tackled.
    \item We provide a case study on Relation Classification in the scientific domain, to fill a gap on cross-domain and cross-dataset evaluation.
\end{itemize}

\section{Relation Extraction Datasets Survey}
\label{sec:survey}

RE has been broadly studied in the last decades and many datasets were published. 
We survey widely used RE datasets in chronological order, and broadly classify them into three domains based on the data source: (1) news and web, (2) scientific publications and (3) Wikipedia.
An overview of the datasets is given in Table~\ref{tab:relatedworks}. 
Our empirical target here focuses on the scientific domain as so far it has received no attention in the cross-domain direction; a similar investigation on overlaps in data, annotation, and model transferability between datasets in other domains is interesting future work.  

\begin{table*}
\centering
\small
\begin{threeparttable}
\begin{tabular}{lllr}
\toprule
\textbf{Dataset} & \textbf{Paper} & \textbf{Data Source} & \textbf{\# Relation Types} \\
\midrule
\multicolumn{3}{l}{\textbf{News and Web}} \\
\midrule
CoNLL04 & \citet{roth-yih-2004-linear} & News articles & 5 \\
ACE\tnote{$\star$} & \citet{doddington-etal-2004-automatic} & News and conversations & 24 \\
NYT & \citet{NYT} & New York Times articles & 24-57\tnote{$\diamond$} \\
SemEval-2007 & \citet{girju-etal-2007-semeval} & Sentences from the web & 7 \\
SemEval-2010 & \citet{hendrickx-etal-2010-semeval} & Sentences from the web & 10 \\
TACRED & \citet{zhang-etal-2017-position} & Newswire and web text & 42 \\
FSL TACRED &\citet{few-shot-tacred} & TACRED data & 42 \\
DWIE & \citet{ZAPOROJETS2021102563} & Deutsche Welle articles & 65 \\
\midrule
\multicolumn{3}{l}{\textbf{Scientific publications}} \\
\midrule
ScienceIE & \citet{augenstein-etal-2017-semeval} &  Scientific articles & 2 \\
SemEval-2018 & \citet{gabor-etal-2018-semeval} & NLP abstracts & 6 \\
{\scshape SciERC} & \citet{luan-etal-2018-multi} & Abstracts of AI proceedings & 7 \\
\midrule
\multicolumn{3}{l}{\textbf{Wikipedia}} \\
\midrule
GoogleRE & - & Wikipedia & 5 \\
mLAMA\tnote{$\star$} & \citet{kassner-etal-2021-multilingual} & GoogleRE data & 5 \\
FewRel & \citet{han-etal-2018-fewrel} & Wikipedia & 100 \\
FewRel 2.0 & \citet{gao-etal-2019-fewrel} & FewRel data + Biomedical literature & 100 + 25 \\
DocRED & \citet{yao-etal-2019-docred} & Wikipedia and Wikidata & 96 \\
SMiLER & \citet{seganti-etal-2021-multilingual} & Wikipedia & 36 \\
\bottomrule
\end{tabular}
\caption{
\label{tab:relatedworks}
Overview of the RE datasets for the English language grouped by macro domains. ($\star$): Multilingual datasets. ($\diamond$): The original paper does not state the number of considered relations and different work describe different dataset setups.
}
\end{threeparttable}
\end{table*}

The CoNLL 2004 dataset~\cite{roth-yih-2004-linear} is one of the first works. It contains annotations for named entities and relations in news articles.
In the same year, the widely studied ACE dataset was published by~\citet{doddington-etal-2004-automatic}.
It contains annotated entities, relations and events in broadcast transcripts, newswire and newspaper data in English, Chinese and Arabic. The corpus is divided into six domains.

Another widely used dataset is The New York Times (NYT) Annotated Corpus,\footnote{\scriptsize{\url{http://iesl.cs.umass.edu/riedel/ecml/}}} first presented by~\citet{NYT}.
It contains over 1.8 million articles by the NYT between 1987 and 2007.
NYT has been created with a distant supervision approach~\cite{mintz-etal-2009-distant}, using Freebase~\cite{bollacker2008freebase} as knowledge base.
Two further versions of it followed recently:
\citet{zhu-etal-2020-towards-understanding} (NYT-H) and~\citet{jia-etal-2019-arnor} published manually annotated versions of the test set in order to perform a more accurate evaluation.

RE has also been part of the SemEval shared tasks for four times so far. The two early SemEval shared tasks focused on the identification of semantic relations between nominals~\cite{nastase2021semantic}. 
For SemEval-2007 Task 4, \citet{girju-etal-2007-semeval} released a dataset for RC into seven generic semantic relations between nominals.
Three years later, for SemEval-2010 Task 8, \citet{hendrickx-etal-2010-semeval} revised the annotation guidelines and published a corpus for RC, by providing  a much larger dataset (10k instances, in comparison to 1.5k of the 2007 shared task).

Since 2017, three RE datasets in the scientific domain emerged, two of the three as SemEval shared tasks.
In SemEval-2017 Task 10~\citet{augenstein-etal-2017-semeval} proposed a dataset for the identification of keyphrases and considered two generic relations (\texttt{HYPONYM-OF} and \texttt{SYNONYM-OF}).
The dataset is called ScienceIE and consists of 500 journal articles from the Computer Science, Material Sciences and Physics fields.
The year after, \citet{gabor-etal-2018-semeval} proposed a corpus for RC and RE made of abstracts of scientific papers from the ACL Anthology for SemEval-2018 Task 7.
The data will be described in further detail in Section~\ref{datasets}. Following the same line, \citet{luan-etal-2018-multi} published \scierc, which is a scientific RE dataset further annotated for coreference resolution.
It contains abstracts from scientific AI-related conferences.
From the existing three scientific RE datasets summarized in Table~\ref{tab:relatedworks}, in our empirical investigation we focus on two (SemEval-2018 and \scierc{}). We leave out ScienceIE as it focuses on keyphrase extraction and it contains two generic relations only.

The Wikipedia domain has been first introduced in 2013. Google released GoogleRE,\footnote{\scriptsize{\url{https://code.google.com/archive/p/relation-extraction-corpus/downloads}}} a RE corpus consisting of snippets from Wikipedia.
More recently, \citet{kassner-etal-2021-multilingual} proposed mLAMA, a multilingual version (53 languages) of GoogleRE with the purpose of investigating knowledge in pre-trained language models.
The multi-lingual dimension is  gaining more interest for RE. Following this trend, \citet{seganti-etal-2021-multilingual} presented SMiLER, a multilingual dataset (14 languages) from Wikipedia with relations belonging to nine domains.

Previous datasets were restricted to the same label collection in the training set and in the test set.
To address this gap and make RE experimental scenarios more realistic, \citet{han-etal-2018-fewrel} published Few-Rel, a Wikipedia-based few-shot learning (FSL) RC dataset annotated by crowdworkers. One year later, \citet{gao-etal-2019-fewrel} published a new version (Few-Rel 2.0), adding a new test set in the biomedical domain and the \texttt{None-Of-The-Above} relation (cf.\ Section~\ref{sec:the-task}).

Back to the news domain, \citet{zhang-etal-2017-position} published a large-scale RE dataset built over newswire and web text, by crowdsourcing relation annotations for sentences with named entity pairs. This resulted in the TACRED dataset with over 100k instances, which is particularly well-suited for neural models.
\citet{few-shot-tacred} used TACRED to make a FSL RC dataset and compared it to FewRel 1.0 and FewRel 2.0, aiming at a more realistic scenario (i.e., non-uniform label distribution, inclusion of pronouns and common nouns).

All datasets so far present a sentence level annotation.
To address this, \citet{yao-etal-2019-docred} published DocRED, a document-level RE dataset from Wikipedia and Wikidata. The difference with a traditional sentence-level corpus is that both the intra- and inter-sentence relations are annotated, increasing the challenge level. In addition to RE, DocRED annotates coreference chains.
DWIE by \citet{ZAPOROJETS2021102563} is another document-level dataset, specifically designed for multi-task IE (Named Entity Recognition, Coreference Resolution,  Relation Extraction, and Entity Linking).

Lastly, there are works focusing on creating datasets for specific RE aspects.
\citet{cheng-etal-2021-hacred}, for example, proposed a Chinese document-level RE dataset for \textit{hard cases} in order to move towards even more challenging evaluation setups.

\paragraph{Domains in RE}
Given our analysis, we observe a shift in target domains: from news text in seminal works, over web texts, to emerging corpora in the scientific domain and the most recent focus on Wikipedia. Similarly, we observe the emerging trend for FSL.

Different datasets lend themselves to study different aspects of the task. Concerning cross-domain RE, we propose to distinguish three setups:
\begin{enumerate}
    \item Data from different domains, but same relation types, which are general enough to be present in each domain (limited and often confined to the ACE dataset)~\cite[e.g.,][]{plank-moschitti-2013-embedding}.
    \item Stable data domain, but different relation sets~\cite[e.g., FewRel by][]{han-etal-2018-fewrel}. Note that when labels change, approaches such as FSL must be adopted.
    \item A combination of both: The data changes and so do the relation types~\cite[e.g., FewRel 2.0 by][]{gao-etal-2019-fewrel}.
\end{enumerate}

In the case study of this paper, given the scientific datasets available, we focus on the first setup.

\section{The Relation Extraction Task}
\label{sec:the-task}

\begin{figure}
\centering
\includegraphics[width=\columnwidth]{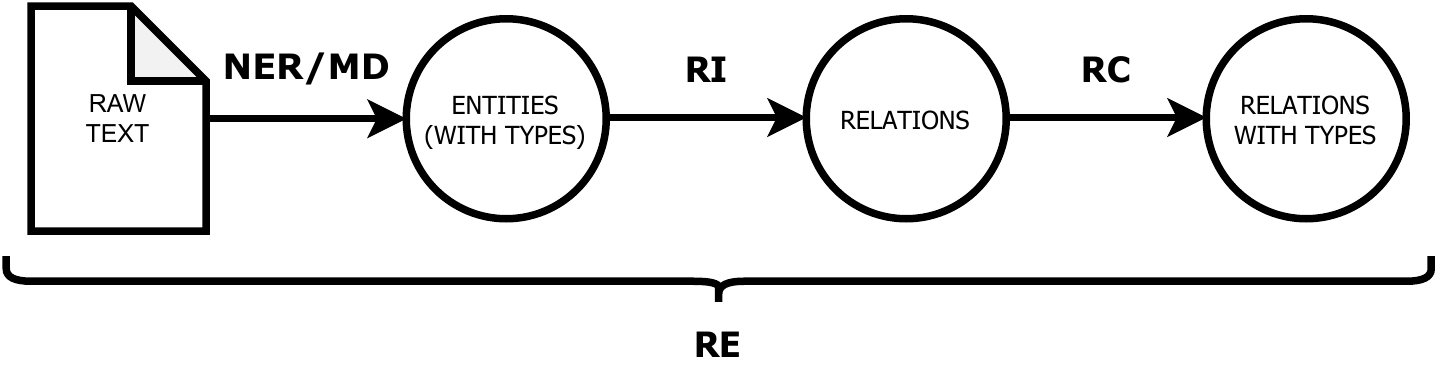}
\caption{Relation Extraction pipeline. NER: Named Entity Recognition; MD: Mention Detection; RI: Relation Identification; RC: Relation Classification.}
\label{fig:RE-task}
\end{figure}

Conceptually, RE involves a pipeline of steps (see Figure~\ref{fig:RE-task}).
Starting from the raw text, the first step consists in identifying the entities and eventually assigning them a type.
Entities involve either nominals or named entities, and hence it is either Named Entity Recognition (NER) or, more broadly, Mention Detection (MD).\footnote{Some studies divide the entity extraction into two sub-steps: identification (often called MD), and subsequent classification into entity types.}
After entities are identified, approaches start to be more blurry as studies have approached RE via different angles.

One way is to take two steps, Relation Identification (RI) and subsequent Relation Classification (RC)~\cite{ye-etal-2019-exploiting}, as illustrated in Figure~\ref{fig:RE-task}.
This means to first identify from all the possible entity pairs the ones which are in some kind of relation via a binary classification task (RI).
As the proportion of positive samples over the negative is usually extremely unbalanced towards the latter~\cite{gormley-etal-2015-improved}, a priori heuristics are generally applied to reduce the possible combinations (e.g., entity pairs involving distant entities, or entity type pairs not licensed by the relations are not even considered).
The last step (RC) is usually a multi-class classification to assign a relation type $r$ to the positive samples from the previous step.
Some studies merge RI and RC~\cite{seganti-etal-2021-multilingual} into one step, by adding a \texttt{no-relation} (\texttt{no-rel}) label. Other studies instead reduce the task to RC, and assume there exists a relation between two entities and the task is to determine the type (without a \texttt{no-rel} label).
Regardless, RI is influenced by the RC setup: Relations which are not in the RC label set are considered as negative samples in the RI phase.
Some studies address this approximation by distinguishing between the \texttt{no-rel} and the \texttt{None-Of-The-Above} (\texttt{NOTA}) relation~\cite{gao-etal-2019-fewrel}. Note that, in our definition, the \texttt{NOTA} label differs from \texttt{no-rel} in the sense that a relation holds between the two entities, but its type is not in the considered RC label set.\footnote{Some studies name such relation \texttt{Other}~\cite{hendrickx-etal-2010-semeval}.}

\paragraph{What Do You Mean by Relation Extraction?} RE studies rarely address the whole pipeline.
We analyze all the ACL papers published in the last five years which contain the \emph{Relation Extraction} keyword in the title and determine which sub-task is performed (NER/MD, RI, RC).
Table~\ref{tab:ACL-RE-history} shows such investigation.
We leave out from this analysis (\textit{a}) papers which make use of distant supervision or which somehow involve knowledge bases, (\textit{b}) shared task papers, (\textit{c}) the bioNLP field, (\textit{d}) temporal RE, and (\textit{e}) Open RE. 

\begin{table}[h]
\resizebox{\columnwidth}{!}{
\centering
\small
\begin{tabular}{l|ccc}
\toprule
\multirow{2}{*}{\textbf{Relation Extraction Paper}} & \multicolumn{3}{c}{\textbf{Task Performed}} \\
 &  \textbf{NER/MD} & \textbf{RI} & \textbf{RC} \\
\midrule
\multicolumn{4}{l}{\textbf{2021}} \\
\midrule
\citet{wang-etal-2021-unire} & \checkmark & \checkmark & \checkmark \\
\citet{cui-etal-2021-refining} &  & & \checkmark \\
\citet{tang-etal-2021-discourse} &  & (?) & \checkmark \\
\citet{xie-etal-2021-revisiting} & \checkmark & (?) & \checkmark \\
\citet{tian-etal-2021-dependency} &  & & \checkmark \\
\citet{ma-etal-2021-sent} &  & \checkmark & \checkmark \\
\citet{mathur-etal-2021-timers} &  & & \checkmark \\
\citet{yang-etal-2021-entity} &  & & \checkmark \\
\citet{huang-etal-2021-three} &  & (?) & \checkmark \\
\citet{huang-etal-2021-entity} &  & (?) & \checkmark \\
\midrule
\multicolumn{4}{l}{\textbf{2020}} \\
\midrule
\citet{kruiper-etal-2020-laymans} & \checkmark &  & \checkmark \\
\citet{nan-etal-2020-reasoning} &  &  & \checkmark \\
\citet{alt-etal-2020-tacred} &  & \checkmark & \checkmark \\
\citet{yu-etal-2020-dialogue} &  & \checkmark & \checkmark \\
\citet{shahbazi-etal-2020-relation} &  & (?) & \checkmark \\
\citet{pouran-ben-veyseh-etal-2020-exploiting} &  &  & \checkmark \\
\midrule
\multicolumn{4}{l}{\textbf{2019}} \\
\midrule
\citet{trisedya-etal-2019-neural} & \checkmark & (?) & \checkmark \\
\citet{guo-etal-2019-attention} &  & \checkmark & \checkmark \\
\citet{yao-etal-2019-docred} &  &  & \checkmark \\
\citet{zhu-etal-2019-graph} &  & \checkmark & \checkmark \\
\citet{li-etal-2019-entity} &  \checkmark  & (?) & \checkmark \\
\citet{ye-etal-2019-exploiting} &  & \checkmark & \checkmark \\
\citet{fu-etal-2019-graphrel} & \checkmark & \checkmark & \checkmark \\
\citet{dixit-al-onaizan-2019-span} & \checkmark & \checkmark & \checkmark \\
\citet{obamuyide-vlachos-2019-meta} &  & (?) & \checkmark \\
\midrule
\multicolumn{4}{l}{\textbf{2018}} \\
\midrule
\citet{christopoulou-etal-2018-walk} &  & \checkmark & \checkmark \\
\citet{phi-etal-2018-ranking} &  &  & \checkmark \\
\midrule
\multicolumn{4}{l}{\textbf{2017}} \\
\midrule
\citet{lin-etal-2017-neural} &  & (?) & \checkmark \\
\bottomrule
\end{tabular}
}
\caption{\label{tab:ACL-RE-history}
ACL paper analysis:\ over the last 5 years, which RE sub-task is performed. \emph{(?)} indicates that either the paper does not state if the step is considered, either it is performed, but in a simplified scenario.
}
\end{table}

The result shows that gold entities are usually assumed for RE, presumably given the complexity of the NER/MD task on its own. Most importantly, for end-to-end models, recent work has shown that ablations for steps like NER are lacking~\citep{taille-etal-2020-lets}.
Our analysis further shows that it is difficult to determine the RI setup. While RC is always performed, the situation is different for RI (or \texttt{no-rel}).
Sometimes RI is clearly not done (i.e., the paper assumes a scenario in which every instance contains at least one relation), but most of the times it is either not clear from the paper, or done in a simplified scenario (e.g., datasets which already clear out most of the \texttt{no-rel} entity pair instances). As this blurriness hampers fair evaluation, we propose that \textit{studies clearly state which step they include}, i.e., whether the work focus is on RC, RI+RC or the full RE pipeline and how special cases (\texttt{no-rel} and \texttt{NOTA}) are handled. These details are utterly important as they impact both model estimation and evaluation.

\paragraph{Pipeline or Joint Model?}
The traditional RE pipeline is, by definition of pipeline, prone to error propagation by sub-tasks.
Joint entity and relation extraction approaches have been proposed in order to alleviate this problem~\citep{miwa-bansal-2016-end,zhang-etal-2017-end,bekoulis-etal-2018-adversarial,BEKOULIS201834,wang-lu-2020-two,wang-etal-2021-unire}.
However, \citet{taille-etal-2020-lets} recently discussed the challenge of properly evaluating such complex models.
They surveyed the evaluation metrics of recently published works on end-to-end RE referring to the \emph{Strict}, \emph{Boundaries}, \emph{Relaxed} evaluation setting proposed by~\citet{bekoulis-etal-2018-adversarial}.
They observe unfair comparisons and overestimations of end-to-end models, and claim the need for more rigorous reports of evaluation settings, including detailed datasets statistics.

While some recent work shifts to joint models, it is still an open question which approach (joint or pipeline) is the most robust.~\citet{zhong-chen-2021-frustratingly} found that when incorporating modern pre-trained language models (e.g., BERT) using separate encoders can surpass existing joint models.
Since the output label space is different, separate encoders could better capture distinct contextual information.
At the moment it is not clear if one approach is more suitable than the other for RE.
For this reason and because of our final goal, which is a closer look to sub-domains in the scientific field, we follow the pipeline approach and, following most work from Table~\ref{tab:ACL-RE-history}, we here restrict the setup by focusing on the RC task.

\paragraph{Open Issues}
To summarize, open issues are: \textit{1)} The unclarity of RE setups, as illustrated in Table~\ref{tab:ACL-RE-history} ---specially regarding RI---leads to problematic evaluation comparisons; \textit{2)} A lack of cross-domain studies, for all three setups outlined in Section~\ref{sec:survey}.

\section{Scientific Domain Data Analysis}
\label{sec:scientific-study}

In this section, we present the two English corpora involved in the experimental study (Section~\ref{datasets}), explain the label mapping adopted for the cross-dataset experiments (Section~\ref{sec:cross-data}), discuss the overlap between the datasets and the annotation divergence between them (Section~\ref{sec:annotation-disagreement}), and introduce the sub-domains considered (Section~\ref{sec:exp-sub-domains}).

\subsection{Datasets}
\label{datasets}

\paragraph{SemEval-2018 Task 7~\cite{gabor-etal-2018-semeval}}
The corpus contains 500 abstracts of published research papers in computational linguistics from the ACL Anthology.
Relations are classified into six classes.
The task was split into three sub-tasks: (1.1) RC on clean data (manually annotated), (1.2)  RC on noisy data (automatically annotated entities) and (2) RI+RC (identifying instances + assigning class labels).
For each sub-task, the training data contains 350 abstracts and the test data 150.
The train set for sub-task (1.1) and (2) is identical.

\paragraph{\scierc{}~\cite{luan-etal-2018-multi}}
The dataset consists of 500 abstracts from scientific publications annotated for entities, their relations and coreference clusters.
The authors define six scientific entity types and seven relation types.
The original paper presents a unified multi-task model for entity extraction, RI+RC and coreference resolution.
\scierc{} is assembled from different conference proceedings. As the data is released with original abstract IDs, this allows us to identify four major sub-domains: AI and ML, Computer Vision (CV), Speech Processing, and NLP, sampled over a time frame from 1980 to 2016.
Details of the sub-domains are provided in Table~\ref{tab:SciERC-domains} in Appendix~\ref{sec:appendix-conferences}. To the best of our knowledge, we are the first to analyze the corpus at this sub-domain level.

\subsection{Cross-dataset Label Mapping}
\label{sec:cross-data}

We homogenize the relation label sets via a manual analysis performed after an exploratory data analysis, as we find that most of the labels in SemEval-2018 and \scierc{} have a direct correspondent, and hence we mapped them as shown in Table~\ref{tab:gold-labels}.
The gold label distribution of the relations on the two datasets is shown in Figure~\ref{fig:gold-label-distribution} in Appendix~\ref{sec:appendix-gold-label-distribution}.
We decided to leave out the two generic labels from {\scshape SciERC} and one relation from SemEval-2018 which does not have any correspondent and is rare.

\begin{table}
\centering
\small
\resizebox{\columnwidth}{!}{
\begin{threeparttable}
\begin{tabular}{l|lr}
\toprule
& \textbf{SemEval-2018} & \textbf{{\scshape SciERC}}\\
\midrule
\multirow{5}{*}{\shortstack[l]{Considered in\\this study}} & 
\texttt{COMPARE} & \texttt{COMPARE} \\
& \texttt{USAGE} & \texttt{USED-FOR} \\
& \texttt{PART\_WHOLE} & \texttt{PART-OF} \\
& \texttt{MODEL-FEATURE} & \texttt{FEATURE-OF} \\
& \texttt{RESULT} & \texttt{EVALUATE-FOR}\tnote{*} \\
\midrule
\multirow{3}{*}{Not-considered} & 
\texttt{TOPIC} & - \\
& - & \texttt{HYPONYM-OF} \\
& - & \texttt{CONJUNCTION} \\
\bottomrule
\end{tabular}
\caption{\label{tab:gold-labels}
Label mapping. (*):\ Same semantic relation, but inverse direction.
We homogenized the two versions by flipping the head with the tail.
}
\end{threeparttable}
}
\end{table}

\subsection{Overlap of the Datasets and Annotation Divergences}
\label{sec:annotation-disagreement}

Our analysis further reveals a high overlap in articles between SemEval-2018 and \scierc{} corresponding to 307 ACL abstracts.\footnote{Note that in our study, regarding SemEval-2018, for fair comparison with {\scshape SciERC}, which is manually annotated, we consider the dataset related to sub-task (1.1).}
Interestingly, the overlap contains a huge annotation divergence.
In more detail, we identify three main annotation disagreement scenarios between the two datasets (represented by the 3 samples in Table~\ref{tab:annotation-disagreement}):

\begin{itemize}
    \item \textbf{Sample 1}: \emph{The annotated entities differ and so the annotated relations do as well.} SemEval-2018 annotates just one entity and thus there can not even exist a relation; as the corresponding sentence in \scierc{} is annotated with two entities, it contains a relation.
    \item \textbf{Sample 2}: \emph{The amount of annotated entities and the amount of annotated relations are the same, but the annotations do not match.}
    The relations involve non-mutual entities and so do not correspond.
    \item \textbf{Sample 3}: \emph{The annotated entities are the same, but the relation annotations differ.} This involves conflicting annotations, e.g., the bold arrow shows the same entity pair annotated with a different relation label.
\end{itemize}

Table~\ref{tab:datasets-annotation} shows the annotation statistics from the two corpora and their overlap.
Overall both datasets contain the same amount of abstracts, but the amount of annotated relations differs substantially.
The overlap between the two corpora reports a similar trend.
Even the fairer count of the common labels (see Table~\ref{tab:gold-labels}) reveals that the annotation gap still holds (ratio of 1:1.8).
In more detail, the entity pairs annotated in both dataset by using a strict criterion (i.e., entity spans with the same boundaries) are only 394 (considering relations from the whole relation sets).
Out of them, only 327 are labeled with the same relation type, meaning that there are 67 conflicting instances as the bold arrow in Table~\ref{tab:annotation-disagreement} (Sample 3).

\begin{table}
\centering
\small
\begin{tabular}{lcc}
\toprule
\multicolumn{3}{l}{\textbf{Whole corpus}} \\
\midrule
& SemEval-2018 & {\scshape SciERC} \\
\midrule
\# abstracts & 500 & 500 \\
\# relations & 1,583 & 4,648 \\
\midrule
\multicolumn{3}{l}{\textbf{Datasets Overlap} (307 abstracts)} \\
\midrule
\# relations & 1,087 & 2,476 \\
\# common relations & 1,071 & 1,922 \\
\midrule
\multicolumn{2}{l}{Same entity pair} & 394 \\
\multicolumn{2}{l}{Same entity pair + same relation type} & 327 \\
\bottomrule
\end{tabular}
\caption{\label{tab:datasets-annotation}
SemEval-2018 and {\scshape SciERC} annotation comparison. The common relations are the ones with a direct correspondent in both datasets (see Table~\ref{tab:gold-labels}).
}
\end{table}

\begin{table*}[ht]
\resizebox{\textwidth}{!}{
\centering
\begin{tabular}{l|l}
\toprule
\multicolumn{2}{l}{\textbf{Sample 1: Different number of entity (and relation) annotations}} \\
\hline
SemEval-2018 &
\begin{dependency}
\begin{deptext}
We evaluate the utility of this \underline{constraint} in two different algorithms.\\
\end{deptext}
\end{dependency}
\\
 {\scshape SciERC} &
\begin{dependency}
\begin{deptext}
We evaluate the utility of this \& \underline{constraint} \& in two different \& \underline{algorithms} \&. \\
\end{deptext}
\depedge[edge height=2ex]{4}{2}{EVALUATE-FOR}
\end{dependency}
\\
\midrule
\multicolumn{2}{l}{\textbf{Sample 2: Different entity annotations}} \\
\hline
SemEval-2018 &
\begin{dependency}
\begin{deptext}
We propose a \underline{detection method} for orthographic variants caused by  \& \underline{transliteration}  \& in a large  \& \underline{corpus} \&.\\
\end{deptext}
\depedge[edge height=2ex]{2}{4}{PART\_WHOLE}
\end{dependency}
\\
{\scshape SciERC} &
\begin{dependency}
\begin{deptext}
We propose a \& \underline{detection method} \& for \& \underline{orthographic variants} \& caused by \underline{transliteration} in a large corpus. \\
\end{deptext}
\depedge[edge height=2ex]{2}{4}{USED-FOR}
\end{dependency}
\\
\midrule
\multicolumn{2}{l}{\textbf{Sample 3: Different relation annotations}} \\
\hline
SemEval-2018 &
\begin{dependency}[edge height=2ex]
\begin{deptext}
The \underline{speech-search algorithm} is implemented on a \& \underline{board} \& with a single \& \underline{Intel i860 chip} \&, which provides a factor of 5 speed-up over a
\underline{SUN 4} for \underline{straight C code}\&.\\
\end{deptext}
\depedge[edge style={ultra thick}, label style={font=\bfseries,thick}]{4}{2}{MODEL-FEATURE}
\end{dependency}
\\
\begin{tabular}{l}
{\scshape SciERC}
\end{tabular} &
\begin{dependency}[edge height=2ex]
\begin{deptext}
The \& \underline{speech-search algorithm} \& is implemented on a \& \underline{board} \& with a single \& \underline{Intel i860 chip} \&, which provides a factor of 5 speed-up over a \&
\underline{SUN 4} \& for \& \underline{straight C code}\&. \\
\end{deptext}
\depedge{2}{4}{USED-FOR}
\depedge[edge style={ultra thick}, label style={font=\bfseries,thick}]{6}{4}{PART-OF}
\depedge{6}{8}{COMPARE}
\depedge[edge height=4.5ex]{6}{10}{USED-FOR}
\depedge{8}{10}{USED-FOR}
\end{dependency}
\\
\bottomrule
\end{tabular}
}
\caption{\label{tab:annotation-disagreement}
Annotated sentence pairs from SemEval-2018 and \scierc{}. The underlined spans are the entities.
}
\end{table*}

\subsection{Experimental Sub-domains}
\label{sec:exp-sub-domains}

We use the metadata described in Section~\ref{datasets} to divide {\scshape SciERC} into four sub-domains.
Figure~\ref{fig:gold-label-distribution-domains} in Appendix~\ref{sec:appendix-gold-label-distribution} shows the label distribution over the new {\scshape SciERC} split.
As we are particularly interested in the annotation divergence impact, we leave out of this study 193 abstracts from SemEval-2018 which are not in overlap with \scierc{}.

\begin{table}
\centering
\small
\begin{tabular}{llrrr}
\toprule
\textbf{Dataset} & \textbf{Sub-domain} & \textbf{train} & \textbf{dev} & \textbf{test} \\
\midrule
SemEval-2018 & NLP & 257 & 50 & 50 \\
\midrule
\multirow{4}{*}{\scierc} & NLP & 257 & 50 & 50  \\
& AI-ML & - & - & 52 \\
& CV & - & - & 105 \\
& SPEECH & - & - & 35 \\
\bottomrule
\end{tabular}
\caption{\label{tab:final-domains}
Sub-domains and relative amount of abstracts.
}
\end{table}

We assume a setup which takes the NLP domain as source training domain in all experiments, as it is the largest sub-domain in both datasets.
The considered sub-domains and their relative amount of data are reported in Table~\ref{tab:final-domains}.

\section{Experiments}
\label{sec:experiments}

\subsection{Model Setup}

Since the seminal work by~\citet{nguyen-grishman-2015-relation}, Convolutional Neural Networks (CNNs) are widely used for IE tasks~\citep{zeng-etal-2014-relation,nguyen-grishman-2015-relation,fu-etal-2017-domain,augenstein-etal-2017-semeval,gabor-etal-2018-semeval,yao-etal-2019-docred}. Similarly, since the advent of contextualized representations~\citep{peters-etal-2018-deep,devlin-etal-2019-bert}, BERT-like representations are commonly used~\citep{seganti-etal-2021-multilingual}, but non-contextualized embeddings (i.e., GloVe, fastText) are still widely adopted~\citep{yao-etal-2019-docred,huang-etal-2021-three}. We compare the best CNN setup to fine-tuning a full transformer model. For the latter we use the MaChAmp toolkit~\cite{van-der-goot-etal-2021-massive}

Our CNN follows~\citet{nguyen-grishman-2015-relation}. We tests both non-contextualized word embeddings---fastText ~\cite{bojanowski-etal-2017-enriching}---and contextualized ones---BERT~\cite{devlin-etal-2019-bert} and the domain-specific SciBERT~\cite{beltagy-etal-2019-scibert}.
Further details about the model implementation and hyperparameter settings can be found in Appendix~\ref{sec:appendix-model}.
We use macro F1-score as evaluation metric. All experiments were run over three different seeds and the results reported are the mean.
\footnote{\scriptsize{\href{https://github.com/elisabassignana/scientific-re}{\url{https://github.com/elisabassignana/scientific-re}}}}

\subsection{Cross-dataset Evaluation}
\label{sec:cross-domain-exp}

We test the following training configurations:\footnote{The development set follows the train set distributions.}
(1) \emph{cross-dataset}: Training on SemEval-2018 and testing on \scierc{}, and vice versa; (2) \emph{cross-annotation}: Training on a mix of SemEval-2018 and \scierc{} overlap: (2.1) \emph{exclusive}: Considering either abstracts from the two corpora, (2.2) \emph{repeated labeling}: Including every abstract twice, once from each dataset; this approach repeats instances with different annotations and is a simple method to handle divergences in annotation~\citep{sheng2008,survey_disagreement}, (2.3) \emph{filter}: Double annotation of the abstracts as in (2.2), but filtering out conflicting annotations.

\paragraph{Results}
Table~\ref{tab:results} reports the results of the experiments.
The \emph{cross-dataset} experiments (1) confirm the expected drop across datasets, in both directions (Sem: 40.28 $\rightarrow$ 34.81 and {\scshape Sci}: 34.29 $\rightarrow$ 31.37).
Considering the \emph{cross-annotation} setups, results are mixed in the \emph{exclusive} version (2.1). The overall amount of training data is the same as the cross-dataset experiments, but there is less dataset-specific data, which hurts SemEval-2018.
In contrast, regarding (2.2) and (2.3), in both setups improvements are evident on both test sets.
Compared to (2.1), the training data amount is effectively doubled and the model benefits from it. Removing the conflicting instances results in a slightly smaller train set, but an overall higher average performance (43.81 $\rightarrow$ 44.16).
The improvement of (2.3) over (2.2) is significant, which we test by the \emph{almost stochastic dominance} test~\cite{dror-etal-2019-deep}. Details about significance are in Appendix~\ref{sec:appendix-significance}.

\begin{table*}
\resizebox{\textwidth}{!}{
\centering
\small
\begin{tabular}{l|ccccc|cc|cc}
\toprule
\textbf{Model} & \multicolumn{7}{c|}{CNN} & \multicolumn{2}{c}{Transformer [tuned]} \\
\midrule
\textbf{Word embedding} & \multicolumn{5}{c|}{FastText} & BERT & SciBERT & SciBERT & SciBERT \\
\midrule
$\downarrow$\textbf{Test} | \textbf{Train} (NLP) $\rightarrow$ & Sem & {\scshape Sci} & $[$\sfrac{1}{2} + \sfrac{1}{2}$]$ & 2A & 2A w/o CR & 2A w/o CR & 2A w/o CR & 2A & 2A w/o CR \\
\midrule
SemEval NLP & 40.28 & 34.81 & 39.91 & 50.17 & 48.95 & 42.54 & 49.27 & 79.16 & 77.79 \\
\scierc{} NLP & 31.37 & 34.29 & 36.29 & 39.36 & 41.48 & 38.63 & 51.99 & 67.36 & 69.90 \\
\scierc{} AI-ML & 37.00 & 50.44 & 46.78 & 49.52 & 49.66 & 40.81 & 51.14 & 72.48 & 76.80 \\
\scierc{} CV & 33.32 & 41.30 & 37.24 & 44.59 & 45.60 & 38.51 & 48.18 & 73.55 & 76.11 \\
\scierc{} SPEECH & 29.60 & 35.00 & 33.71 & 35.39 & 35.11 & 31.62 & 42.72 & 64.17 & 65.21 \\
\midrule
\textbf{avg.} & 34.31 & 39.17 & 38.79 & 43.81 & \textbf{44.16} & 38.34 & \textbf{48.66} & 71.34 & \textbf{73.56} \\
\bottomrule
\end{tabular}
}
\caption{
\label{tab:results}
Macro F1-scores of the cross-dataset and cross-domain experiments.
(2.1) $[$\sfrac{1}{2} + \sfrac{1}{2}$]$ refers to the case in which the train is made half by SemEval-2018 and half by {\scshape SciERC}; (2.2) 2A means double annotation from the two datasets; (2.3) CR are the conflicting relations (bold sample in Table~\ref{tab:annotation-disagreement}).
}
\end{table*}

\subsection{Contextualized Word Embeddings}

We pick the best performing training scenario (\emph{cross-annotation filter}, 2.3) and compare fastText with contextualized embeddings: BERT and the domain-specific SciBERT.
The central columns of Table~\ref{tab:results} report the results.
While BERT does not bring relevant improvements over the best fastText setup, SciBERT confirms the strength of domain-specific trained language models (improvement of 4.5 F1 points and \emph{almost stochastic dominance}). Compared to the CNN, full transformer fine-tuning results in the best model (rightmost columns). We tested different setups to feed the input to the transformer (see appendix~\ref{sec:appendix-machamp}), finding two entity spans and the full sentence as best setup.
The full fine-tuned transformer model confirms the \textit{dominance} of training setup (2.3) over (2.2).

\subsection{Cross-domain Evaluation}
\label{sec:sub-domain-analysis}
Next, we look at  \emph{cross-domain} variation: Training on NLP, and testing on all sub-domains. The lower rows in Table~\ref{tab:results} show the results.
If we focus on the SciBERT models, we observe that there is some drop in performance from NLP, but mostly to CV and SPEECH.
Interestingly, in some cases, AI-ML even outperforms the in-domain performance.
Over all models, the SPEECH domain shows the clearest drop in transfer from NLP.\footnote{We note that the data amount for speech is the smallest in respect to the other sub-corpora, which might have an impact.}
From an analysis of the predictions of the RC trained on SciBERT, we notice that the classifier struggles with identifying the most frequent \texttt{USAGE} relation (see Appendix~\ref{sec:appendix-gold-label-distribution}) across sub-domains 
(confusion from lowest to highest:  AI-ML, CV and SPEECH), 
and it is most confused with \texttt{MODEL-FEATURE}.
Figure~\ref{fig:cm-sub-domains} in Appendix~\ref{sec:appendix-pca} contains the detailed confusion matrices.
The overall evaluation suggests that in this setup sub-domain variation impacts RC performance to a limiting degree only.

\begin{figure}
\centering
\includegraphics[width=\columnwidth]{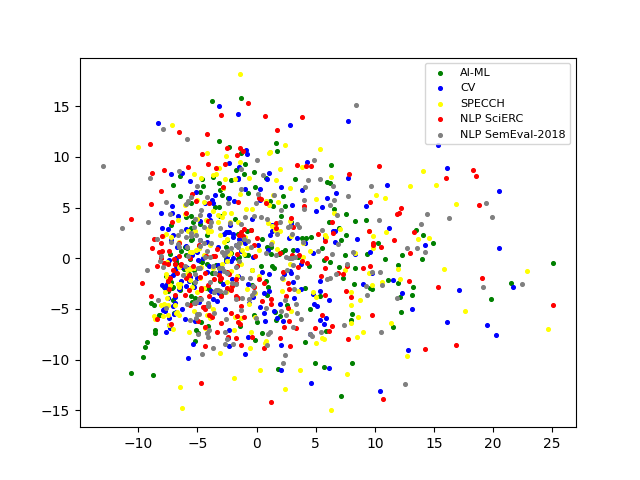}
\caption{PCA representation of the CNN hidden state (just before the linear layer) using SciBERT.}
\label{fig:pca-scibert}
\end{figure}

In order to confirm this qualitatively, we (1) inspect whether model-internal representations are able to capture sub-domain variation, and we (2) test whether sub-domain variation is identifiable. To answer (1),  we visualise the PCA representation of the CNN trained on setup (2.3) with SciBERT. The result is shown in Figure~\ref{fig:pca-scibert}. The plot confirms that the representations do not contain visible clusters: The relation instances from each sub-domain are equally spread over it, and thus the performance of the relation classifier is similar for each of them.
Our intuition is that the unified label set contains relations general enough to be equally covered by every sub-domain.

We explore the sub-domains more deeply apart from the RC task.
To answer (2), we built a domain classifier to investigate how hard it is to tear apart the sub-domains. We hypothesize that, if sub-domains are distinguishable, a classifier should be able to easily distinguish them by looking at the coarsest level (the abstract). 
The classifier consists of a linear layer on top of the SciBERT encoder and achieves a F1-score of 62.01, over a random baseline of 25.58. This shows that the sub-domains are identifiable at the abstract level but with modest performance.
As we would expect, SPEECH and NLP are highly confused (Figure~\ref{fig:cm-conf-class} in Appendix~\ref{sec:appendix-cm} reports the confusion matrix) and the large vocabulary overlap shown in Table~\ref{tab:vocabulary-overlap} between these sub-domains confirms this observation. Overall, sub-domains are identifiable but have limited impact on the RC task in the setup considered.

\begin{table}
\small
\centering
\begin{tabular}{lccc}
\toprule
\textbf{Domain} & \textbf{\# word types} & \textbf{\# overlap} & \textbf{\% overlap} \\
\midrule
NLP & 5,646 & - & - \\
AI-ML & 1,895 & 917 & 48.39\% \\
CV & 3,387 & 1,205 & 35.58\% \\
SPEECH & 1,398 & 715 & 51.14\% \\
\bottomrule
\end{tabular}
\caption{\label{tab:vocabulary-overlap}
Vocabulary overlap between NLP and the other sub-domains. \emph{\# word types}, \emph{\# overlap}  in word types, and \emph{\% overlap} as relative percentages.
Note that the amount of abstracts varies, cf.\ Table~\ref{tab:final-domains}.
}
\end{table}

\section{Conclusions}

We present a survey on datasets for
RE, revisit the task definition, and provide an empirical study on scientific RC.
We observe a domain shift in RE datasets, and a trend towards multilingual and 
FSL for RE.
Our analysis shows that our surveyed ACL RE papers focus mostly on RC and assume gold entities. Other steps are more blurry, concluding with a call for reporting RE setups more clearly.

As testing on only one dataset or domain bears risks of overestimation, we carry out a cross-dataset evaluation. Despite large data overlaps, we find annotations to substantially differ, which impacts classification results.
Sub-domains extracted from meta-data instead only slightly impact performance.
This finding on sub-domain variation is specific to the explored RC task on the scientific setup considered.
Our study contributes to the first of three cross-domain RE setups we propose (Section~\ref{sec:survey}) to aid further work on  generalization for RE.

\section*{Limitations and Ethical Considerations}

This work focuses on a limited view of the whole RE  research field.
Our dataset survey excludes specific angles of RE such as temporal RE or bioNLP, as they are large sub-fields which warrant a dedicated analysis in itself.
From a methodological point of view, in our analysis we did not further cover  weakly-supervised (e.g., distant supervision) and un-supervised approaches.
Finally, given that our study points out gaps in RE, specifically cross-dataset, our experiments are still limited to RC only and next steps are to extend to the whole pipeline and to additional datasets and domains. 

The data analyzed in this work is based on existing publicly-available datasets  (based on published research papers).

\section*{Acknowledgements}
We thank the NLPnorth group for insightful discussions on this work---in particular Mike Zhang and Max Müller-Eberstein. We would also like to thank the anonymous reviewers for their comments to improve this paper. Last, we also thank the ITU's High-performance Computing cluster for computing resources.
This research is supported by the Independent Research Fund Denmark (Danmarks Frie Forskningsfond; DFF) grant number 9063-00077B.

\bibliography{anthology,custom}

\begin{thebibliography}{88}
\expandafter\ifx\csname natexlab\endcsname\relax\def\natexlab#1{#1}\fi

\bibitem[{Alt et~al.(2020)Alt, Gabryszak, and Hennig}]{alt-etal-2020-tacred}
Christoph Alt, Aleksandra Gabryszak, and Leonhard Hennig. 2020.
\newblock \href {https://doi.org/10.18653/v1/2020.acl-main.142} {{TACRED}
  revisited: A thorough evaluation of the {TACRED} relation extraction task}.
\newblock In \emph{Proceedings of the 58th Annual Meeting of the Association
  for Computational Linguistics}, pages 1558--1569, Online. Association for
  Computational Linguistics.

\bibitem[{Augenstein et~al.(2017)Augenstein, Das, Riedel, Vikraman, and
  McCallum}]{augenstein-etal-2017-semeval}
Isabelle Augenstein, Mrinal Das, Sebastian Riedel, Lakshmi Vikraman, and Andrew
  McCallum. 2017.
\newblock \href {https://doi.org/10.18653/v1/S17-2091} {{S}em{E}val 2017 task
  10: {S}cience{IE} - extracting keyphrases and relations from scientific
  publications}.
\newblock In \emph{Proceedings of the 11th International Workshop on Semantic
  Evaluation ({S}em{E}val-2017)}, pages 546--555, Vancouver, Canada.
  Association for Computational Linguistics.

\bibitem[{Aydar et~al.(2021)Aydar, Bozal, and {\"O}zbay}]{aydar2021neural}
Mehmet Aydar, {\"O}zge Bozal, and Furkan {\"O}zbay. 2021.
\newblock Neural relation extraction: a review.
\newblock \emph{Turkish Journal of Electrical Engineering \& Computer
  Sciences}, 29(2):1029--1043.

\bibitem[{Bach and Badaskar(2007)}]{bach2007review}
Nguyen Bach and Sameer Badaskar. 2007.
\newblock A review of relation extraction.
\newblock \emph{Literature review for Language and Statistics II}, 2:1--15.

\bibitem[{Baldini~Soares et~al.(2019)Baldini~Soares, FitzGerald, Ling, and
  Kwiatkowski}]{baldini-soares-etal-2019-matching}
Livio Baldini~Soares, Nicholas FitzGerald, Jeffrey Ling, and Tom Kwiatkowski.
  2019.
\newblock \href {https://doi.org/10.18653/v1/P19-1279} {Matching the blanks:
  Distributional similarity for relation learning}.
\newblock In \emph{Proceedings of the 57th Annual Meeting of the Association
  for Computational Linguistics}, pages 2895--2905, Florence, Italy.
  Association for Computational Linguistics.

\bibitem[{Bekoulis et~al.(2018{\natexlab{a}})Bekoulis, Deleu, Demeester, and
  Develder}]{bekoulis-etal-2018-adversarial}
Giannis Bekoulis, Johannes Deleu, Thomas Demeester, and Chris Develder.
  2018{\natexlab{a}}.
\newblock \href {https://doi.org/10.18653/v1/D18-1307} {Adversarial training
  for multi-context joint entity and relation extraction}.
\newblock In \emph{Proceedings of the 2018 Conference on Empirical Methods in
  Natural Language Processing}, pages 2830--2836, Brussels, Belgium.
  Association for Computational Linguistics.

\bibitem[{Bekoulis et~al.(2018{\natexlab{b}})Bekoulis, Deleu, Demeester, and
  Develder}]{BEKOULIS201834}
Giannis Bekoulis, Johannes Deleu, Thomas Demeester, and Chris Develder.
  2018{\natexlab{b}}.
\newblock \href {https://doi.org/https://doi.org/10.1016/j.eswa.2018.07.032}
  {Joint entity recognition and relation extraction as a multi-head selection
  problem}.
\newblock \emph{Expert Systems with Applications}, 114:34--45.

\bibitem[{Beltagy et~al.(2019)Beltagy, Lo, and
  Cohan}]{beltagy-etal-2019-scibert}
Iz~Beltagy, Kyle Lo, and Arman Cohan. 2019.
\newblock \href {https://doi.org/10.18653/v1/D19-1371} {{S}ci{BERT}: A
  pretrained language model for scientific text}.
\newblock In \emph{Proceedings of the 2019 Conference on Empirical Methods in
  Natural Language Processing and the 9th International Joint Conference on
  Natural Language Processing (EMNLP-IJCNLP)}, pages 3615--3620, Hong Kong,
  China. Association for Computational Linguistics.

\bibitem[{Bojanowski et~al.(2017)Bojanowski, Grave, Joulin, and
  Mikolov}]{bojanowski-etal-2017-enriching}
Piotr Bojanowski, Edouard Grave, Armand Joulin, and Tomas Mikolov. 2017.
\newblock \href {https://doi.org/10.1162/tacl_a_00051} {Enriching word vectors
  with subword information}.
\newblock \emph{Transactions of the Association for Computational Linguistics},
  5:135--146.

\bibitem[{Bollacker et~al.(2008)Bollacker, Evans, Paritosh, Sturge, and
  Taylor}]{bollacker2008freebase}
Kurt Bollacker, Colin Evans, Praveen Paritosh, Tim Sturge, and Jamie Taylor.
  2008.
\newblock Freebase: a collaboratively created graph database for structuring
  human knowledge.
\newblock In \emph{Proceedings of the 2008 ACM SIGMOD international conference
  on Management of data}, pages 1247--1250.

\bibitem[{Bonferroni(1936)}]{bonferroni1936teoria}
C.E. Bonferroni. 1936.
\newblock \href {https://books.google.de/books?id=3CY-HQAACAAJ} {\emph{Teoria
  statistica delle classi e calcolo delle probabilit{\`a}}}.
\newblock Pubblicazioni del R. Istituto superiore di scienze economiche e
  commerciali di Firenze. Seeber.

\bibitem[{Cheng et~al.(2021)Cheng, Liu, Qu, Zhao, Liang, Wang, Huai, Yuan, and
  Xiao}]{cheng-etal-2021-hacred}
Qiao Cheng, Juntao Liu, Xiaoye Qu, Jin Zhao, Jiaqing Liang, Zhefeng Wang,
  Baoxing Huai, Nicholas~Jing Yuan, and Yanghua Xiao. 2021.
\newblock \href {https://doi.org/10.18653/v1/2021.findings-acl.249}
  {{H}ac{RED}: A large-scale relation extraction dataset toward hard cases in
  practical applications}.
\newblock In \emph{Findings of the Association for Computational Linguistics:
  ACL-IJCNLP 2021}, pages 2819--2831, Online. Association for Computational
  Linguistics.

\bibitem[{Christopoulou et~al.(2018)Christopoulou, Miwa, and
  Ananiadou}]{christopoulou-etal-2018-walk}
Fenia Christopoulou, Makoto Miwa, and Sophia Ananiadou. 2018.
\newblock \href {https://doi.org/10.18653/v1/P18-2014} {A walk-based model on
  entity graphs for relation extraction}.
\newblock In \emph{Proceedings of the 56th Annual Meeting of the Association
  for Computational Linguistics (Volume 2: Short Papers)}, pages 81--88,
  Melbourne, Australia. Association for Computational Linguistics.

\bibitem[{Cui et~al.(2021)Cui, Yang, Yu, Hu, Cheng, Yi, and
  Xiao}]{cui-etal-2021-refining}
Li~Cui, Deqing Yang, Jiaxin Yu, Chengwei Hu, Jiayang Cheng, Jingjie Yi, and
  Yanghua Xiao. 2021.
\newblock \href {https://doi.org/10.18653/v1/2021.acl-long.20} {Refining sample
  embeddings with relation prototypes to enhance continual relation
  extraction}.
\newblock In \emph{Proceedings of the 59th Annual Meeting of the Association
  for Computational Linguistics and the 11th International Joint Conference on
  Natural Language Processing (Volume 1: Long Papers)}, pages 232--243, Online.
  Association for Computational Linguistics.

\bibitem[{Devlin et~al.(2019)Devlin, Chang, Lee, and
  Toutanova}]{devlin-etal-2019-bert}
Jacob Devlin, Ming-Wei Chang, Kenton Lee, and Kristina Toutanova. 2019.
\newblock \href {https://doi.org/10.18653/v1/N19-1423} {{BERT}: Pre-training of
  deep bidirectional transformers for language understanding}.
\newblock In \emph{Proceedings of the 2019 Conference of the North {A}merican
  Chapter of the Association for Computational Linguistics: Human Language
  Technologies, Volume 1 (Long and Short Papers)}, pages 4171--4186,
  Minneapolis, Minnesota. Association for Computational Linguistics.

\bibitem[{Dixit and Al-Onaizan(2019)}]{dixit-al-onaizan-2019-span}
Kalpit Dixit and Yaser Al-Onaizan. 2019.
\newblock \href {https://doi.org/10.18653/v1/P19-1525} {Span-level model for
  relation extraction}.
\newblock In \emph{Proceedings of the 57th Annual Meeting of the Association
  for Computational Linguistics}, pages 5308--5314, Florence, Italy.
  Association for Computational Linguistics.

\bibitem[{Doddington et~al.(2004)Doddington, Mitchell, Przybocki, Ramshaw,
  Strassel, and Weischedel}]{doddington-etal-2004-automatic}
George Doddington, Alexis Mitchell, Mark Przybocki, Lance Ramshaw, Stephanie
  Strassel, and Ralph Weischedel. 2004.
\newblock \href {http://www.lrec-conf.org/proceedings/lrec2004/pdf/5.pdf} {The
  automatic content extraction ({ACE}) program {--} tasks, data, and
  evaluation}.
\newblock In \emph{Proceedings of the Fourth International Conference on
  Language Resources and Evaluation ({LREC}{'}04)}, Lisbon, Portugal. European
  Language Resources Association (ELRA).

\bibitem[{Dror et~al.(2019)Dror, Shlomov, and Reichart}]{dror-etal-2019-deep}
Rotem Dror, Segev Shlomov, and Roi Reichart. 2019.
\newblock \href {https://doi.org/10.18653/v1/P19-1266} {Deep dominance - how to
  properly compare deep neural models}.
\newblock In \emph{Proceedings of the 57th Annual Meeting of the Association
  for Computational Linguistics}, pages 2773--2785, Florence, Italy.
  Association for Computational Linguistics.

\bibitem[{Fu et~al.(2017)Fu, Nguyen, Min, and Grishman}]{fu-etal-2017-domain}
Lisheng Fu, Thien~Huu Nguyen, Bonan Min, and Ralph Grishman. 2017.
\newblock \href {https://aclanthology.org/I17-2072} {Domain adaptation for
  relation extraction with domain adversarial neural network}.
\newblock In \emph{Proceedings of the Eighth International Joint Conference on
  Natural Language Processing (Volume 2: Short Papers)}, pages 425--429,
  Taipei, Taiwan. Asian Federation of Natural Language Processing.

\bibitem[{Fu et~al.(2019)Fu, Li, and Ma}]{fu-etal-2019-graphrel}
Tsu-Jui Fu, Peng-Hsuan Li, and Wei-Yun Ma. 2019.
\newblock \href {https://doi.org/10.18653/v1/P19-1136} {{G}raph{R}el: Modeling
  text as relational graphs for joint entity and relation extraction}.
\newblock In \emph{Proceedings of the 57th Annual Meeting of the Association
  for Computational Linguistics}, pages 1409--1418, Florence, Italy.
  Association for Computational Linguistics.

\bibitem[{G{\'a}bor et~al.(2018)G{\'a}bor, Buscaldi, Schumann, QasemiZadeh,
  Zargayouna, and Charnois}]{gabor-etal-2018-semeval}
Kata G{\'a}bor, Davide Buscaldi, Anne-Kathrin Schumann, Behrang QasemiZadeh,
  Ha{\"\i}fa Zargayouna, and Thierry Charnois. 2018.
\newblock \href {https://doi.org/10.18653/v1/S18-1111} {{S}em{E}val-2018 task
  7: Semantic relation extraction and classification in scientific papers}.
\newblock In \emph{Proceedings of The 12th International Workshop on Semantic
  Evaluation}, pages 679--688, New Orleans, Louisiana. Association for
  Computational Linguistics.

\bibitem[{Gao et~al.(2019)Gao, Han, Zhu, Liu, Li, Sun, and
  Zhou}]{gao-etal-2019-fewrel}
Tianyu Gao, Xu~Han, Hao Zhu, Zhiyuan Liu, Peng Li, Maosong Sun, and Jie Zhou.
  2019.
\newblock \href {https://doi.org/10.18653/v1/D19-1649} {{F}ew{R}el 2.0: Towards
  more challenging few-shot relation classification}.
\newblock In \emph{Proceedings of the 2019 Conference on Empirical Methods in
  Natural Language Processing and the 9th International Joint Conference on
  Natural Language Processing (EMNLP-IJCNLP)}, pages 6250--6255, Hong Kong,
  China. Association for Computational Linguistics.

\bibitem[{Girju et~al.(2007)Girju, Nakov, Nastase, Szpakowicz, Turney, and
  Yuret}]{girju-etal-2007-semeval}
Roxana Girju, Preslav Nakov, Vivi Nastase, Stan Szpakowicz, Peter Turney, and
  Deniz Yuret. 2007.
\newblock \href {https://aclanthology.org/S07-1003} {{S}em{E}val-2007 task 04:
  Classification of semantic relations between nominals}.
\newblock In \emph{Proceedings of the Fourth International Workshop on Semantic
  Evaluations ({S}em{E}val-2007)}, pages 13--18, Prague, Czech Republic.
  Association for Computational Linguistics.

\bibitem[{Gorman and Bedrick(2019)}]{gorman-bedrick-2019-need}
Kyle Gorman and Steven Bedrick. 2019.
\newblock \href {https://doi.org/10.18653/v1/P19-1267} {We need to talk about
  standard splits}.
\newblock In \emph{Proceedings of the 57th Annual Meeting of the Association
  for Computational Linguistics}, pages 2786--2791, Florence, Italy.
  Association for Computational Linguistics.

\bibitem[{Gormley et~al.(2015)Gormley, Yu, and
  Dredze}]{gormley-etal-2015-improved}
Matthew~R. Gormley, Mo~Yu, and Mark Dredze. 2015.
\newblock \href {https://doi.org/10.18653/v1/D15-1205} {Improved relation
  extraction with feature-rich compositional embedding models}.
\newblock In \emph{Proceedings of the 2015 Conference on Empirical Methods in
  Natural Language Processing}, pages 1774--1784, Lisbon, Portugal. Association
  for Computational Linguistics.

\bibitem[{Grishman(2012)}]{grishman2012information}
Ralph Grishman. 2012.
\newblock Information extraction: Capabilities and challenges.

\bibitem[{Guo et~al.(2019)Guo, Zhang, and Lu}]{guo-etal-2019-attention}
Zhijiang Guo, Yan Zhang, and Wei Lu. 2019.
\newblock \href {https://doi.org/10.18653/v1/P19-1024} {Attention guided graph
  convolutional networks for relation extraction}.
\newblock In \emph{Proceedings of the 57th Annual Meeting of the Association
  for Computational Linguistics}, pages 241--251, Florence, Italy. Association
  for Computational Linguistics.

\bibitem[{Han et~al.(2018)Han, Zhu, Yu, Wang, Yao, Liu, and
  Sun}]{han-etal-2018-fewrel}
Xu~Han, Hao Zhu, Pengfei Yu, Ziyun Wang, Yuan Yao, Zhiyuan Liu, and Maosong
  Sun. 2018.
\newblock \href {https://doi.org/10.18653/v1/D18-1514} {{F}ew{R}el: A
  large-scale supervised few-shot relation classification dataset with
  state-of-the-art evaluation}.
\newblock In \emph{Proceedings of the 2018 Conference on Empirical Methods in
  Natural Language Processing}, pages 4803--4809, Brussels, Belgium.
  Association for Computational Linguistics.

\bibitem[{Hendrickx et~al.(2010)Hendrickx, Kim, Kozareva, Nakov,
  {\'O}~S{\'e}aghdha, Pad{\'o}, Pennacchiotti, Romano, and
  Szpakowicz}]{hendrickx-etal-2010-semeval}
Iris Hendrickx, Su~Nam Kim, Zornitsa Kozareva, Preslav Nakov, Diarmuid
  {\'O}~S{\'e}aghdha, Sebastian Pad{\'o}, Marco Pennacchiotti, Lorenza Romano,
  and Stan Szpakowicz. 2010.
\newblock \href {https://aclanthology.org/S10-1006} {{S}em{E}val-2010 task 8:
  Multi-way classification of semantic relations between pairs of nominals}.
\newblock In \emph{Proceedings of the 5th International Workshop on Semantic
  Evaluation}, pages 33--38, Uppsala, Sweden. Association for Computational
  Linguistics.

\bibitem[{Huang et~al.(2021{\natexlab{a}})Huang, Qi, Wang, Ma, and
  Huang}]{huang-etal-2021-entity}
Kevin Huang, Peng Qi, Guangtao Wang, Tengyu Ma, and Jing Huang.
  2021{\natexlab{a}}.
\newblock \href {https://doi.org/10.18653/v1/2021.repl4nlp-1.30} {Entity and
  evidence guided document-level relation extraction}.
\newblock In \emph{Proceedings of the 6th Workshop on Representation Learning
  for NLP (RepL4NLP-2021)}, pages 307--315, Online. Association for
  Computational Linguistics.

\bibitem[{Huang et~al.(2021{\natexlab{b}})Huang, Zhu, Feng, Ye, Lai, and
  Zhao}]{huang-etal-2021-three}
Quzhe Huang, Shengqi Zhu, Yansong Feng, Yuan Ye, Yuxuan Lai, and Dongyan Zhao.
  2021{\natexlab{b}}.
\newblock \href {https://doi.org/10.18653/v1/2021.acl-short.126} {Three
  sentences are all you need: Local path enhanced document relation
  extraction}.
\newblock In \emph{Proceedings of the 59th Annual Meeting of the Association
  for Computational Linguistics and the 11th International Joint Conference on
  Natural Language Processing (Volume 2: Short Papers)}, pages 998--1004,
  Online. Association for Computational Linguistics.

\bibitem[{Jia et~al.(2019)Jia, Dai, Xiao, and Wu}]{jia-etal-2019-arnor}
Wei Jia, Dai Dai, Xinyan Xiao, and Hua Wu. 2019.
\newblock \href {https://doi.org/10.18653/v1/P19-1135} {{ARNOR}: Attention
  regularization based noise reduction for distant supervision relation
  classification}.
\newblock In \emph{Proceedings of the 57th Annual Meeting of the Association
  for Computational Linguistics}, pages 1399--1408, Florence, Italy.
  Association for Computational Linguistics.

\bibitem[{Kassner et~al.(2021)Kassner, Dufter, and
  Sch{\"u}tze}]{kassner-etal-2021-multilingual}
Nora Kassner, Philipp Dufter, and Hinrich Sch{\"u}tze. 2021.
\newblock \href {https://aclanthology.org/2021.eacl-main.284} {Multilingual
  {LAMA}: Investigating knowledge in multilingual pretrained language models}.
\newblock In \emph{Proceedings of the 16th Conference of the European Chapter
  of the Association for Computational Linguistics: Main Volume}, pages
  3250--3258, Online. Association for Computational Linguistics.

\bibitem[{Kruiper et~al.(2020)Kruiper, Vincent, Chen-Burger, Desmulliez, and
  Konstas}]{kruiper-etal-2020-laymans}
Ruben Kruiper, Julian Vincent, Jessica Chen-Burger, Marc Desmulliez, and
  Ioannis Konstas. 2020.
\newblock \href {https://doi.org/10.18653/v1/2020.acl-main.137} {In layman{'}s
  terms: Semi-open relation extraction from scientific texts}.
\newblock In \emph{Proceedings of the 58th Annual Meeting of the Association
  for Computational Linguistics}, pages 1489--1500, Online. Association for
  Computational Linguistics.

\bibitem[{Li et~al.(2019)Li, Yin, Sun, Li, Yuan, Chai, Zhou, and
  Li}]{li-etal-2019-entity}
Xiaoya Li, Fan Yin, Zijun Sun, Xiayu Li, Arianna Yuan, Duo Chai, Mingxin Zhou,
  and Jiwei Li. 2019.
\newblock \href {https://doi.org/10.18653/v1/P19-1129} {Entity-relation
  extraction as multi-turn question answering}.
\newblock In \emph{Proceedings of the 57th Annual Meeting of the Association
  for Computational Linguistics}, pages 1340--1350, Florence, Italy.
  Association for Computational Linguistics.

\bibitem[{Lin et~al.(2017)Lin, Liu, and Sun}]{lin-etal-2017-neural}
Yankai Lin, Zhiyuan Liu, and Maosong Sun. 2017.
\newblock \href {https://doi.org/10.18653/v1/P17-1004} {Neural relation
  extraction with multi-lingual attention}.
\newblock In \emph{Proceedings of the 55th Annual Meeting of the Association
  for Computational Linguistics (Volume 1: Long Papers)}, pages 34--43,
  Vancouver, Canada. Association for Computational Linguistics.

\bibitem[{Lippincott et~al.(2010)Lippincott, {\'O}~S{\'e}aghdha, Sun, and
  Korhonen}]{lippincott-etal-2010-exploring}
Tom Lippincott, Diarmuid {\'O}~S{\'e}aghdha, Lin Sun, and Anna Korhonen. 2010.
\newblock \href {https://aclanthology.org/C10-1078} {Exploring variation across
  biomedical subdomains}.
\newblock In \emph{Proceedings of the 23rd International Conference on
  Computational Linguistics (Coling 2010)}, pages 689--697, Beijing, China.
  Coling 2010 Organizing Committee.

\bibitem[{Liu(2020)}]{liu_survey_2020}
Kang Liu. 2020.
\newblock \href {https://doi.org/10.1007/s11431-020-1673-6} {A survey on neural
  relation extraction}.
\newblock \emph{Science China Technological Sciences}, 63(10):1971--1989.

\bibitem[{Liu et~al.(2021)Liu, Xu, Yu, Dai, Ji, Cahyawijaya, Madotto, and
  Fung}]{liu2021crossner}
Zihan Liu, Yan Xu, Tiezheng Yu, Wenliang Dai, Ziwei Ji, Samuel Cahyawijaya,
  Andrea Madotto, and Pascale Fung. 2021.
\newblock \href {https://ojs.aaai.org/index.php/AAAI/article/view/17587}
  {Crossner: Evaluating cross-domain named entity recognition}.
\newblock \emph{Proceedings of the AAAI Conference on Artificial Intelligence},
  35(15):13452--13460.

\bibitem[{Luan et~al.(2018)Luan, He, Ostendorf, and
  Hajishirzi}]{luan-etal-2018-multi}
Yi~Luan, Luheng He, Mari Ostendorf, and Hannaneh Hajishirzi. 2018.
\newblock \href {https://doi.org/10.18653/v1/D18-1360} {Multi-task
  identification of entities, relations, and coreference for scientific
  knowledge graph construction}.
\newblock In \emph{Proceedings of the 2018 Conference on Empirical Methods in
  Natural Language Processing}, pages 3219--3232, Brussels, Belgium.
  Association for Computational Linguistics.

\bibitem[{Luo et~al.(2016)Luo, Uzuner, and Szolovits}]{bio-survey}
Yuan Luo, Özlem Uzuner, and Peter Szolovits. 2016.
\newblock \href {https://doi.org/10.1093/bib/bbw001} {{Bridging semantics and
  syntax with graph algorithms—state-of-the-art of extracting biomedical
  relations}}.
\newblock \emph{Briefings in Bioinformatics}, 18(1):160--178.

\bibitem[{Ma et~al.(2021)Ma, Gui, Li, Zhang, Huang, and
  Zhou}]{ma-etal-2021-sent}
Ruotian Ma, Tao Gui, Linyang Li, Qi~Zhang, Xuanjing Huang, and Yaqian Zhou.
  2021.
\newblock \href {https://doi.org/10.18653/v1/2021.acl-long.484} {{SENT}:
  {S}entence-level distant relation extraction via negative training}.
\newblock In \emph{Proceedings of the 59th Annual Meeting of the Association
  for Computational Linguistics and the 11th International Joint Conference on
  Natural Language Processing (Volume 1: Long Papers)}, pages 6201--6213,
  Online. Association for Computational Linguistics.

\bibitem[{Mathur et~al.(2021)Mathur, Jain, Dernoncourt, Morariu, Tran, and
  Manocha}]{mathur-etal-2021-timers}
Puneet Mathur, Rajiv Jain, Franck Dernoncourt, Vlad Morariu, Quan~Hung Tran,
  and Dinesh Manocha. 2021.
\newblock \href {https://doi.org/10.18653/v1/2021.acl-short.67} {{TIMERS}:
  Document-level temporal relation extraction}.
\newblock In \emph{Proceedings of the 59th Annual Meeting of the Association
  for Computational Linguistics and the 11th International Joint Conference on
  Natural Language Processing (Volume 2: Short Papers)}, pages 524--533,
  Online. Association for Computational Linguistics.

\bibitem[{Mintz et~al.(2009)Mintz, Bills, Snow, and
  Jurafsky}]{mintz-etal-2009-distant}
Mike Mintz, Steven Bills, Rion Snow, and Daniel Jurafsky. 2009.
\newblock \href {https://aclanthology.org/P09-1113} {Distant supervision for
  relation extraction without labeled data}.
\newblock In \emph{Proceedings of the Joint Conference of the 47th Annual
  Meeting of the {ACL} and the 4th International Joint Conference on Natural
  Language Processing of the {AFNLP}}, pages 1003--1011, Suntec, Singapore.
  Association for Computational Linguistics.

\bibitem[{Miwa and Bansal(2016)}]{miwa-bansal-2016-end}
Makoto Miwa and Mohit Bansal. 2016.
\newblock \href {https://doi.org/10.18653/v1/P16-1105} {End-to-end relation
  extraction using {LSTM}s on sequences and tree structures}.
\newblock In \emph{Proceedings of the 54th Annual Meeting of the Association
  for Computational Linguistics (Volume 1: Long Papers)}, pages 1105--1116,
  Berlin, Germany. Association for Computational Linguistics.

\bibitem[{Nan et~al.(2020)Nan, Guo, Sekulic, and Lu}]{nan-etal-2020-reasoning}
Guoshun Nan, Zhijiang Guo, Ivan Sekulic, and Wei Lu. 2020.
\newblock \href {https://doi.org/10.18653/v1/2020.acl-main.141} {Reasoning with
  latent structure refinement for document-level relation extraction}.
\newblock In \emph{Proceedings of the 58th Annual Meeting of the Association
  for Computational Linguistics}, pages 1546--1557, Online. Association for
  Computational Linguistics.

\bibitem[{Nastase et~al.(2021)Nastase, Szpakowicz, Nakov, and
  S{\'e}agdha}]{nastase2021semantic}
Vivi Nastase, Stan Szpakowicz, Preslav Nakov, and Diarmuid~{\'O} S{\'e}agdha.
  2021.
\newblock Semantic relations between nominals.
\newblock \emph{Synthesis Lectures on Human Language Technologies},
  14(1):1--234.

\bibitem[{Nguyen and Grishman(2015)}]{nguyen-grishman-2015-relation}
Thien~Huu Nguyen and Ralph Grishman. 2015.
\newblock \href {https://doi.org/10.3115/v1/W15-1506} {Relation extraction:
  Perspective from convolutional neural networks}.
\newblock In \emph{Proceedings of the 1st Workshop on Vector Space Modeling for
  Natural Language Processing}, pages 39--48, Denver, Colorado. Association for
  Computational Linguistics.

\bibitem[{Obamuyide and Vlachos(2019)}]{obamuyide-vlachos-2019-meta}
Abiola Obamuyide and Andreas Vlachos. 2019.
\newblock \href {https://doi.org/10.18653/v1/W19-4326} {Meta-learning improves
  lifelong relation extraction}.
\newblock In \emph{Proceedings of the 4th Workshop on Representation Learning
  for NLP (RepL4NLP-2019)}, pages 224--229, Florence, Italy. Association for
  Computational Linguistics.

\bibitem[{Pawar et~al.(2017)Pawar, Palshikar, and
  Bhattacharyya}]{pawar2017relation}
Sachin Pawar, Girish~K Palshikar, and Pushpak Bhattacharyya. 2017.
\newblock \href {https://arxiv.org/abs/1712.05191} {Relation extraction: A
  survey}.
\newblock \emph{arXiv preprint arXiv:1712.05191}.

\bibitem[{Peters et~al.(2018)Peters, Neumann, Iyyer, Gardner, Clark, Lee, and
  Zettlemoyer}]{peters-etal-2018-deep}
Matthew~E. Peters, Mark Neumann, Mohit Iyyer, Matt Gardner, Christopher Clark,
  Kenton Lee, and Luke Zettlemoyer. 2018.
\newblock \href {https://doi.org/10.18653/v1/N18-1202} {Deep contextualized
  word representations}.
\newblock In \emph{Proceedings of the 2018 Conference of the North {A}merican
  Chapter of the Association for Computational Linguistics: Human Language
  Technologies, Volume 1 (Long Papers)}, pages 2227--2237, New Orleans,
  Louisiana. Association for Computational Linguistics.

\bibitem[{Petrov and McDonald(2012)}]{petrov2012overview}
Slav Petrov and Ryan McDonald. 2012.
\newblock Overview of the 2012 shared task on parsing the web.
\newblock In \emph{First Workshop on Syntactic Analysis of Non-Canonical
  Language (SANCL), NAACL-HLT}.

\bibitem[{Phi et~al.(2018)Phi, Santoso, Shimbo, and
  Matsumoto}]{phi-etal-2018-ranking}
Van-Thuy Phi, Joan Santoso, Masashi Shimbo, and Yuji Matsumoto. 2018.
\newblock \href {https://doi.org/10.18653/v1/P18-2015} {Ranking-based automatic
  seed selection and noise reduction for weakly supervised relation
  extraction}.
\newblock In \emph{Proceedings of the 56th Annual Meeting of the Association
  for Computational Linguistics (Volume 2: Short Papers)}, pages 89--95,
  Melbourne, Australia. Association for Computational Linguistics.

\bibitem[{Plank and Moschitti(2013)}]{plank-moschitti-2013-embedding}
Barbara Plank and Alessandro Moschitti. 2013.
\newblock \href {https://aclanthology.org/P13-1147} {Embedding semantic
  similarity in tree kernels for domain adaptation of relation extraction}.
\newblock In \emph{Proceedings of the 51st Annual Meeting of the Association
  for Computational Linguistics (Volume 1: Long Papers)}, pages 1498--1507,
  Sofia, Bulgaria. Association for Computational Linguistics.

\bibitem[{Pouran Ben~Veyseh et~al.(2020)Pouran Ben~Veyseh, Dernoncourt, Dou,
  and Nguyen}]{pouran-ben-veyseh-etal-2020-exploiting}
Amir Pouran Ben~Veyseh, Franck Dernoncourt, Dejing Dou, and Thien~Huu Nguyen.
  2020.
\newblock \href {https://doi.org/10.18653/v1/2020.acl-main.715} {Exploiting the
  syntax-model consistency for neural relation extraction}.
\newblock In \emph{Proceedings of the 58th Annual Meeting of the Association
  for Computational Linguistics}, pages 8021--8032, Online. Association for
  Computational Linguistics.

\bibitem[{Pradhan et~al.(2013)Pradhan, Moschitti, Xue, Ng, Bj{\"o}rkelund,
  Uryupina, Zhang, and Zhong}]{pradhan-etal-2013-towards}
Sameer Pradhan, Alessandro Moschitti, Nianwen Xue, Hwee~Tou Ng, Anders
  Bj{\"o}rkelund, Olga Uryupina, Yuchen Zhang, and Zhi Zhong. 2013.
\newblock \href {https://aclanthology.org/W13-3516} {Towards robust linguistic
  analysis using {O}nto{N}otes}.
\newblock In \emph{Proceedings of the Seventeenth Conference on Computational
  Natural Language Learning}, pages 143--152, Sofia, Bulgaria. Association for
  Computational Linguistics.

\bibitem[{Pyysalo et~al.(2008)Pyysalo, S{\~{A}}¦tre, Tsujii, and
  Salakoski}]{pyysalo08why}
S.~Pyysalo, R.~S{\~{A}}¦tre, J.~Tsujii, and T.~Salakoski. 2008.
\newblock \href
  {http://mars.cs.utu.fi/smbm2008/files/smbm2008proceedings/smbmpaper_33.pdf}
  {Why biomedical relation extraction results are incomparable and what to do
  about it}.
\newblock In \emph{Proceedings of the Third International Symposium on Semantic
  Mining in Biomedicine (SMBM 2008)}, pages 149--152.

\bibitem[{Riedel et~al.(2010)Riedel, Yao, and McCallum}]{NYT}
Sebastian Riedel, Limin Yao, and Andrew McCallum. 2010.
\newblock Modeling relations and their mentions without labeled text.
\newblock In \emph{Joint European Conference on Machine Learning and Knowledge
  Discovery in Databases}, pages 148--163, Berlin, Heidelberg. Springer Berlin
  Heidelberg.

\bibitem[{Roth and Yih(2004)}]{roth-yih-2004-linear}
Dan Roth and Wen-tau Yih. 2004.
\newblock \href {https://aclanthology.org/W04-2401} {A linear programming
  formulation for global inference in natural language tasks}.
\newblock In \emph{Proceedings of the Eighth Conference on Computational
  Natural Language Learning ({C}o{NLL}-2004) at {HLT}-{NAACL} 2004}, pages
  1--8, Boston, Massachusetts, USA. Association for Computational Linguistics.

\bibitem[{Sabo et~al.(2021)Sabo, Elazar, Goldberg, and Dagan}]{few-shot-tacred}
Ofer Sabo, Yanai Elazar, Yoav Goldberg, and Ido Dagan. 2021.
\newblock \href {https://doi.org/10.1162/tacl_a_00392} {{Revisiting Few-shot
  Relation Classification: Evaluation Data and Classification Schemes}}.
\newblock \emph{Transactions of the Association for Computational Linguistics},
  9:691--706.

\bibitem[{Seganti et~al.(2021)Seganti, Firl{\k{a}}g, Skowronska, Sat{\l}awa,
  and Andruszkiewicz}]{seganti-etal-2021-multilingual}
Alessandro Seganti, Klaudia Firl{\k{a}}g, Helena Skowronska, Micha{\l}
  Sat{\l}awa, and Piotr Andruszkiewicz. 2021.
\newblock \href {https://aclanthology.org/2021.eacl-main.166} {Multilingual
  entity and relation extraction dataset and model}.
\newblock In \emph{Proceedings of the 16th Conference of the European Chapter
  of the Association for Computational Linguistics: Main Volume}, pages
  1946--1955, Online. Association for Computational Linguistics.

\bibitem[{Shahbazi et~al.(2020)Shahbazi, Fern, Ghaeini, and
  Tadepalli}]{shahbazi-etal-2020-relation}
Hamed Shahbazi, Xiaoli Fern, Reza Ghaeini, and Prasad Tadepalli. 2020.
\newblock \href {https://doi.org/10.18653/v1/2020.acl-main.579} {Relation
  extraction with explanation}.
\newblock In \emph{Proceedings of the 58th Annual Meeting of the Association
  for Computational Linguistics}, pages 6488--6494, Online. Association for
  Computational Linguistics.

\bibitem[{Sheng et~al.(2008)Sheng, Provost, and Ipeirotis}]{sheng2008}
Victor~S. Sheng, Foster Provost, and Panagiotis~G. Ipeirotis. 2008.
\newblock \href {https://doi.org/10.1145/1401890.1401965} {Get another label?
  improving data quality and data mining using multiple, noisy labelers}.
\newblock In \emph{Proceedings of the 14th ACM SIGKDD International Conference
  on Knowledge Discovery and Data Mining}, KDD ’08, page 614–622, New York,
  NY, USA. Association for Computing Machinery.

\bibitem[{S{\o}gaard et~al.(2021)S{\o}gaard, Ebert, Bastings, and
  Filippova}]{sogaard-etal-2021-need}
Anders S{\o}gaard, Sebastian Ebert, Jasmijn Bastings, and Katja Filippova.
  2021.
\newblock \href {https://aclanthology.org/2021.eacl-main.156} {We need to talk
  about random splits}.
\newblock In \emph{Proceedings of the 16th Conference of the European Chapter
  of the Association for Computational Linguistics: Main Volume}, pages
  1823--1832, Online. Association for Computational Linguistics.

\bibitem[{Taill{\'e} et~al.(2020)Taill{\'e}, Guigue, Scoutheeten, and
  Gallinari}]{taille-etal-2020-lets}
Bruno Taill{\'e}, Vincent Guigue, Geoffrey Scoutheeten, and Patrick Gallinari.
  2020.
\newblock \href {https://doi.org/10.18653/v1/2020.emnlp-main.301} {Let{'}s
  {S}top {I}ncorrect {C}omparisons in {E}nd-to-end {R}elation {E}xtraction!}
\newblock In \emph{Proceedings of the 2020 Conference on Empirical Methods in
  Natural Language Processing (EMNLP)}, pages 3689--3701, Online. Association
  for Computational Linguistics.

\bibitem[{Tang et~al.(2021)Tang, Lin, Liao, Lu, Han, Sun, Xie, and
  Xu}]{tang-etal-2021-discourse}
Jialong Tang, Hongyu Lin, Meng Liao, Yaojie Lu, Xianpei Han, Le~Sun, Weijian
  Xie, and Jin Xu. 2021.
\newblock \href {https://doi.org/10.18653/v1/2021.acl-long.60} {From discourse
  to narrative: Knowledge projection for event relation extraction}.
\newblock In \emph{Proceedings of the 59th Annual Meeting of the Association
  for Computational Linguistics and the 11th International Joint Conference on
  Natural Language Processing (Volume 1: Long Papers)}, pages 732--742, Online.
  Association for Computational Linguistics.

\bibitem[{Tian et~al.(2021)Tian, Chen, Song, and
  Wan}]{tian-etal-2021-dependency}
Yuanhe Tian, Guimin Chen, Yan Song, and Xiang Wan. 2021.
\newblock \href {https://doi.org/10.18653/v1/2021.acl-long.344}
  {Dependency-driven relation extraction with attentive graph convolutional
  networks}.
\newblock In \emph{Proceedings of the 59th Annual Meeting of the Association
  for Computational Linguistics and the 11th International Joint Conference on
  Natural Language Processing (Volume 1: Long Papers)}, pages 4458--4471,
  Online. Association for Computational Linguistics.

\bibitem[{Trisedya et~al.(2019)Trisedya, Weikum, Qi, and
  Zhang}]{trisedya-etal-2019-neural}
Bayu~Distiawan Trisedya, Gerhard Weikum, Jianzhong Qi, and Rui Zhang. 2019.
\newblock \href {https://doi.org/10.18653/v1/P19-1023} {Neural relation
  extraction for knowledge base enrichment}.
\newblock In \emph{Proceedings of the 57th Annual Meeting of the Association
  for Computational Linguistics}, pages 229--240, Florence, Italy. Association
  for Computational Linguistics.

\bibitem[{Ulmer(2021)}]{dennis_ulmer_2021_4638709}
Dennis Ulmer. 2021.
\newblock \href {https://doi.org/10.5281/zenodo.4638709} {{deep-significance:
  Easy and Better Significance Testing for Deep Neural Networks}}.
\newblock Https://github.com/Kaleidophon/deep-significance.

\bibitem[{Uma et~al.(2021)Uma, Fornaciari, Hovy, Paun, Plank, and
  Poesio}]{survey_disagreement}
Alexandra~N. Uma, Tommaso Fornaciari, Dirk Hovy, Silviu Paun, Barbara Plank,
  and Massimo Poesio. 2021.
\newblock Learning from disagreement: A survey.
\newblock \emph{The Journal of Artificial Intelligence Research}, Forthcoming.

\bibitem[{van~der Goot et~al.(2021)van~der Goot, {\"U}st{\"u}n, Ramponi,
  Sharaf, and Plank}]{van-der-goot-etal-2021-massive}
Rob van~der Goot, Ahmet {\"U}st{\"u}n, Alan Ramponi, Ibrahim Sharaf, and
  Barbara Plank. 2021.
\newblock \href {https://aclanthology.org/2021.eacl-demos.22} {Massive choice,
  ample tasks ({M}a{C}h{A}mp): A toolkit for multi-task learning in {NLP}}.
\newblock In \emph{Proceedings of the 16th Conference of the European Chapter
  of the Association for Computational Linguistics: System Demonstrations},
  pages 176--197, Online. Association for Computational Linguistics.

\bibitem[{Wang and Lu(2020)}]{wang-lu-2020-two}
Jue Wang and Wei Lu. 2020.
\newblock \href {https://doi.org/10.18653/v1/2020.emnlp-main.133} {Two are
  better than one: Joint entity and relation extraction with table-sequence
  encoders}.
\newblock In \emph{Proceedings of the 2020 Conference on Empirical Methods in
  Natural Language Processing (EMNLP)}, pages 1706--1721, Online. Association
  for Computational Linguistics.

\bibitem[{Wang et~al.(2021)Wang, Sun, Wu, Zhou, Li, and
  Yan}]{wang-etal-2021-unire}
Yijun Wang, Changzhi Sun, Yuanbin Wu, Hao Zhou, Lei Li, and Junchi Yan. 2021.
\newblock \href {https://doi.org/10.18653/v1/2021.acl-long.19} {{U}ni{RE}: A
  unified label space for entity relation extraction}.
\newblock In \emph{Proceedings of the 59th Annual Meeting of the Association
  for Computational Linguistics and the 11th International Joint Conference on
  Natural Language Processing (Volume 1: Long Papers)}, pages 220--231, Online.
  Association for Computational Linguistics.

\bibitem[{Williams et~al.(2018)Williams, Nangia, and
  Bowman}]{williams-etal-2018-broad}
Adina Williams, Nikita Nangia, and Samuel Bowman. 2018.
\newblock \href {https://doi.org/10.18653/v1/N18-1101} {A broad-coverage
  challenge corpus for sentence understanding through inference}.
\newblock In \emph{Proceedings of the 2018 Conference of the North {A}merican
  Chapter of the Association for Computational Linguistics: Human Language
  Technologies, Volume 1 (Long Papers)}, pages 1112--1122, New Orleans,
  Louisiana. Association for Computational Linguistics.

\bibitem[{Xie et~al.(2021)Xie, Liang, Liu, Huang, Huang, and
  Xiao}]{xie-etal-2021-revisiting}
Chenhao Xie, Jiaqing Liang, Jingping Liu, Chengsong Huang, Wenhao Huang, and
  Yanghua Xiao. 2021.
\newblock \href {https://doi.org/10.18653/v1/2021.acl-long.277} {Revisiting the
  negative data of distantly supervised relation extraction}.
\newblock In \emph{Proceedings of the 59th Annual Meeting of the Association
  for Computational Linguistics and the 11th International Joint Conference on
  Natural Language Processing (Volume 1: Long Papers)}, pages 3572--3581,
  Online. Association for Computational Linguistics.

\bibitem[{Yang et~al.(2021)Yang, Zhang, Niu, Zhao, and
  Pu}]{yang-etal-2021-entity}
Shan Yang, Yongfei Zhang, Guanglin Niu, Qinghua Zhao, and Shiliang Pu. 2021.
\newblock \href {https://doi.org/10.18653/v1/2021.acl-short.124} {Entity
  concept-enhanced few-shot relation extraction}.
\newblock In \emph{Proceedings of the 59th Annual Meeting of the Association
  for Computational Linguistics and the 11th International Joint Conference on
  Natural Language Processing (Volume 2: Short Papers)}, pages 987--991,
  Online. Association for Computational Linguistics.

\bibitem[{Yao et~al.(2019)Yao, Ye, Li, Han, Lin, Liu, Liu, Huang, Zhou, and
  Sun}]{yao-etal-2019-docred}
Yuan Yao, Deming Ye, Peng Li, Xu~Han, Yankai Lin, Zhenghao Liu, Zhiyuan Liu,
  Lixin Huang, Jie Zhou, and Maosong Sun. 2019.
\newblock \href {https://doi.org/10.18653/v1/P19-1074} {{D}oc{RED}: A
  large-scale document-level relation extraction dataset}.
\newblock In \emph{Proceedings of the 57th Annual Meeting of the Association
  for Computational Linguistics}, pages 764--777, Florence, Italy. Association
  for Computational Linguistics.

\bibitem[{Ye et~al.(2019)Ye, Li, Xie, Sheng, Chen, and
  Zhang}]{ye-etal-2019-exploiting}
Wei Ye, Bo~Li, Rui Xie, Zhonghao Sheng, Long Chen, and Shikun Zhang. 2019.
\newblock \href {https://doi.org/10.18653/v1/P19-1130} {Exploiting entity {BIO}
  tag embeddings and multi-task learning for relation extraction with
  imbalanced data}.
\newblock In \emph{Proceedings of the 57th Annual Meeting of the Association
  for Computational Linguistics}, pages 1351--1360, Florence, Italy.
  Association for Computational Linguistics.

\bibitem[{Yu et~al.(2020)Yu, Sun, Cardie, and Yu}]{yu-etal-2020-dialogue}
Dian Yu, Kai Sun, Claire Cardie, and Dong Yu. 2020.
\newblock \href {https://doi.org/10.18653/v1/2020.acl-main.444} {Dialogue-based
  relation extraction}.
\newblock In \emph{Proceedings of the 58th Annual Meeting of the Association
  for Computational Linguistics}, pages 4927--4940, Online. Association for
  Computational Linguistics.

\bibitem[{Yu et~al.(2019)Yu, Zhang, Yasunaga, Tan, Lin, Li, Er, Li, Pang, Chen,
  Ji, Dixit, Proctor, Shim, Kraft, Zhang, Xiong, Socher, and
  Radev}]{yu-etal-2019-sparc}
Tao Yu, Rui Zhang, Michihiro Yasunaga, Yi~Chern Tan, Xi~Victoria Lin, Suyi Li,
  Heyang Er, Irene Li, Bo~Pang, Tao Chen, Emily Ji, Shreya Dixit, David
  Proctor, Sungrok Shim, Jonathan Kraft, Vincent Zhang, Caiming Xiong, Richard
  Socher, and Dragomir Radev. 2019.
\newblock \href {https://doi.org/10.18653/v1/P19-1443} {{SP}ar{C}: Cross-domain
  semantic parsing in context}.
\newblock In \emph{Proceedings of the 57th Annual Meeting of the Association
  for Computational Linguistics}, pages 4511--4523, Florence, Italy.
  Association for Computational Linguistics.

\bibitem[{Zaporojets et~al.(2021)Zaporojets, Deleu, Develder, and
  Demeester}]{ZAPOROJETS2021102563}
Klim Zaporojets, Johannes Deleu, Chris Develder, and Thomas Demeester. 2021.
\newblock \href {https://doi.org/https://doi.org/10.1016/j.ipm.2021.102563}
  {Dwie: An entity-centric dataset for multi-task document-level information
  extraction}.
\newblock \emph{Information Processing \& Management}, 58(4):102563.

\bibitem[{Zeng et~al.(2014)Zeng, Liu, Lai, Zhou, and
  Zhao}]{zeng-etal-2014-relation}
Daojian Zeng, Kang Liu, Siwei Lai, Guangyou Zhou, and Jun Zhao. 2014.
\newblock \href {https://aclanthology.org/C14-1220} {Relation classification
  via convolutional deep neural network}.
\newblock In \emph{Proceedings of {COLING} 2014, the 25th International
  Conference on Computational Linguistics: Technical Papers}, pages 2335--2344,
  Dublin, Ireland. Dublin City University and Association for Computational
  Linguistics.

\bibitem[{Zhang et~al.(2017{\natexlab{a}})Zhang, Zhang, and
  Fu}]{zhang-etal-2017-end}
Meishan Zhang, Yue Zhang, and Guohong Fu. 2017{\natexlab{a}}.
\newblock \href {https://doi.org/10.18653/v1/D17-1182} {End-to-end neural
  relation extraction with global optimization}.
\newblock In \emph{Proceedings of the 2017 Conference on Empirical Methods in
  Natural Language Processing}, pages 1730--1740, Copenhagen, Denmark.
  Association for Computational Linguistics.

\bibitem[{Zhang et~al.(2017{\natexlab{b}})Zhang, Zhong, Chen, Angeli, and
  Manning}]{zhang-etal-2017-position}
Yuhao Zhang, Victor Zhong, Danqi Chen, Gabor Angeli, and Christopher~D.
  Manning. 2017{\natexlab{b}}.
\newblock \href {https://doi.org/10.18653/v1/D17-1004} {Position-aware
  attention and supervised data improve slot filling}.
\newblock In \emph{Proceedings of the 2017 Conference on Empirical Methods in
  Natural Language Processing}, pages 35--45, Copenhagen, Denmark. Association
  for Computational Linguistics.

\bibitem[{Zhong and Chen(2021)}]{zhong-chen-2021-frustratingly}
Zexuan Zhong and Danqi Chen. 2021.
\newblock \href {https://doi.org/10.18653/v1/2021.naacl-main.5} {A
  frustratingly easy approach for entity and relation extraction}.
\newblock In \emph{Proceedings of the 2021 Conference of the North American
  Chapter of the Association for Computational Linguistics: Human Language
  Technologies}, pages 50--61, Online. Association for Computational
  Linguistics.

\bibitem[{Zhu et~al.(2019)Zhu, Lin, Liu, Fu, Chua, and
  Sun}]{zhu-etal-2019-graph}
Hao Zhu, Yankai Lin, Zhiyuan Liu, Jie Fu, Tat-Seng Chua, and Maosong Sun. 2019.
\newblock \href {https://doi.org/10.18653/v1/P19-1128} {Graph neural networks
  with generated parameters for relation extraction}.
\newblock In \emph{Proceedings of the 57th Annual Meeting of the Association
  for Computational Linguistics}, pages 1331--1339, Florence, Italy.
  Association for Computational Linguistics.

\bibitem[{Zhu et~al.(2020{\natexlab{a}})Zhu, Huang, Zhang, Zhu, and
  Huang}]{zhu-etal-2020-crosswoz}
Qi~Zhu, Kaili Huang, Zheng Zhang, Xiaoyan Zhu, and Minlie Huang.
  2020{\natexlab{a}}.
\newblock \href {https://doi.org/10.1162/tacl_a_00314} {{C}ross{WOZ}: A
  large-scale {C}hinese cross-domain task-oriented dialogue dataset}.
\newblock \emph{Transactions of the Association for Computational Linguistics},
  8:281--295.

\bibitem[{Zhu et~al.(2020{\natexlab{b}})Zhu, Wang, Narayana, Sone, Basu, and
  Wang}]{zhu-etal-2020-towards-understanding}
Wanrong Zhu, Xin Wang, Pradyumna Narayana, Kazoo Sone, Sugato Basu, and
  William~Yang Wang. 2020{\natexlab{b}}.
\newblock \href {https://doi.org/10.18653/v1/2020.emnlp-main.708} {Towards
  understanding sample variance in visually grounded language generation:
  Evaluations and observations}.
\newblock In \emph{Proceedings of the 2020 Conference on Empirical Methods in
  Natural Language Processing (EMNLP)}, pages 8806--8811, Online. Association
  for Computational Linguistics.

\end{thebibliography}
\bibliographystyle{acl_natbib}

\clearpage

\appendix

\section*{Appendix}

\section{\scierc{} Conference Division}
\label{sec:appendix-conferences}

The metadata relative to the IDs of the \scierc{} abstracts contains information about the proceedings in which the papers have been published.
We use this information to divide \scierc{} into four sub-domains as shown in Table~\ref{tab:SciERC-domains}.

\begin{table}[h]
\resizebox{\columnwidth}{!}{
\begin{tabular}{lr}
\toprule
\textbf{Conference} & \textbf{\# abs}\\
\midrule
Artificial Intelligence - Machine Learning (AI-ML) & 52 \\
\midrule
NeurIPS & 20 \\
Neural Information Processing Systems \\
IJCAI & 14 \\
International Joint Conference on Artificial Intelligence \\
ICML & 10 \\
International Conference on Machine Learning \\
AAAI & 8 \\
Association for the Advancement of Artificial Intelligence \\
\midrule
Computer Vision (CV) & 105 \\
\midrule
CVPR & 66 \\
Conference on Computer Vision and Pattern Recognition \\
ICCV & 23 \\
International Conference on Computer Vision \\
ECCV & 16 \\
European Conference on Computer Vision \\
\midrule
Speech & 35 \\
\midrule
INTERSPEECH & 25 \\
Annual Conference of the International Speech \\
Communication Association \\
ICASSP & 10 \\
International Conference on Acoustics, Speech, and \\
Signal Processing \\
\midrule
Natural Language Processing (NLP) & 308 \\
\midrule
ACL & 307 \\
Association for Computational Linguistics \\
IJCNLP & 1 \\
International Joint Conference on Natural Language \\
Processing \\
\bottomrule
\end{tabular}
}
\caption{\label{tab:SciERC-domains}
\scierc{} division into conferences and relative amount of abstracts for each of them.
}
\end{table}

\section{Data Analysis}
\label{sec:appendix-gold-label-distribution}

\begin{figure}
\centering
\begin{subfigure}[b]{\columnwidth}
\centering
\caption{SemEval-2018}
\includegraphics[width=\textwidth]{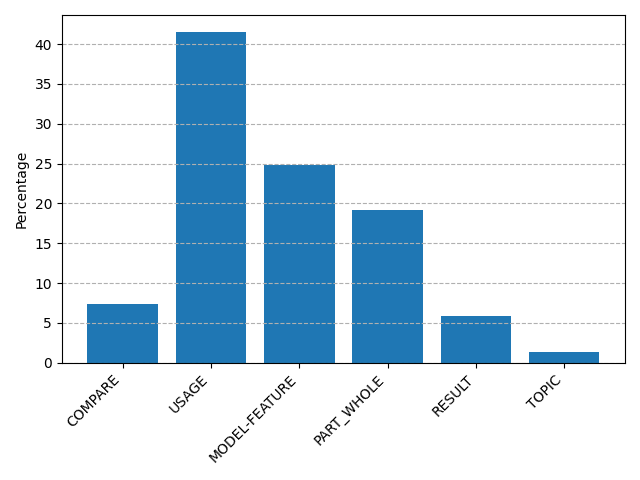}
\label{fig:SemEval-gold-label-distribution}
\end{subfigure}
\begin{subfigure}[b]{\columnwidth}
\centering
\caption{{\scshape SciERC}}
\includegraphics[width=\textwidth]{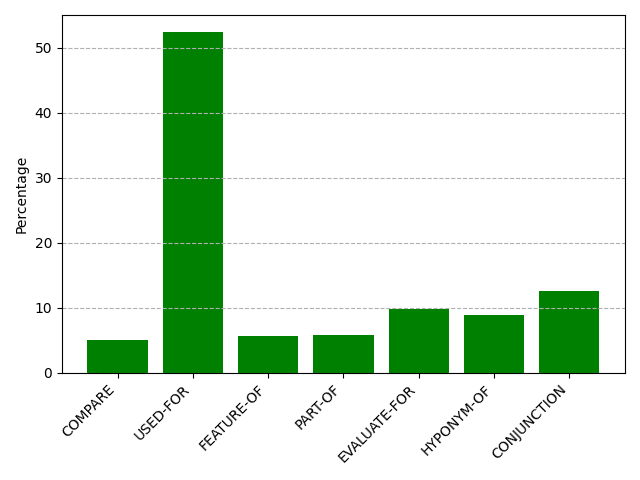}
\label{fig:SciERC-gold-label-distribution}
\end{subfigure}
\caption{Gold label distribution in the SemEval-2018 sub-task (1.1) and {\scshape SciERC} datasets.}
\label{fig:gold-label-distribution}
\end{figure}

Figure~\ref{fig:gold-label-distribution} reports the gold label distribution over SemEval-2018 and \scierc{} respectively.

\begin{figure}
\centering
\includegraphics[width=\columnwidth]{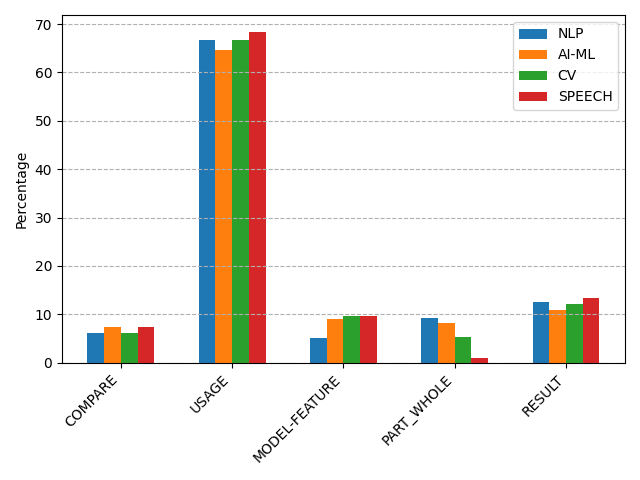}
\caption{Gold label distribution of the five considered relations over {\scshape SciERC} sub-domains.}
\label{fig:gold-label-distribution-domains}
\end{figure}

Figure~\ref{fig:gold-label-distribution-domains}, instead, contains the gold label distributions of \scierc{} sub-domains over the five matching labels between the two datasets (see Table~\ref{tab:gold-labels}).

\section{Model Details}
\label{sec:appendix-model}

Our RC model is a CNN with four layers~\cite{nguyen-grishman-2015-relation}.
The layers consist of lookup embedding layers for word embeddings and entity position information (detailed below), convolutional layers with n-gram kernel sizes (2, 3 and 4), a max-pooling layer and a linear softmax relation classification layer with dropout of 0.5. 
Each input to the network is a sentence containing a pair of entities---which positions in the sentence are given---and a label within $R$, the set of five considered relations.

We experiment with three types of pre-trained \textit{word embeddings}: one non-contextualized, fastText~\cite{bojanowski-etal-2017-enriching}, and two contextualized representations, BERT~\cite{devlin-etal-2019-bert} and the domain-specific SciBERT~\cite{beltagy-etal-2019-scibert}. For word split into subword-tokens, we adopt the strategy of keeping only the first embedding for each token.
For every token we also consider two \textit{position embeddings} following~\citet{nguyen-grishman-2015-relation}. Each of them encodes the relative distance of the token from each of the two entities involved in the relation.

Hyperparameters were determined by tuning the model on a held-out development set.

All experiments were ran on an NVIDIA\textsuperscript{®} A100 SXM4 40 GB GPU and an AMD EPYC\textsuperscript{™} 7662 64-Core CPU.

\section{Significance Testing}
\label{sec:appendix-significance}

We compare our setups using Almost Stochastic Order (ASO;~\citealp{dror-etal-2019-deep}).\footnote{Implementation by~\citet{dennis_ulmer_2021_4638709}.}
Given the results over multiple seeds, the ASO test determines whether there is a stochastic order. 
The method computes a score ($\epsilon_{min}$) which represents how far the first is from being significantly better in respect to the second. The possible scenarios are therefore (\emph{a}) $\epsilon_{min} = 0.0$ (\emph{truly stochastic dominance}) and (\emph{b}) $\epsilon_{min} < 0.5$ (\emph{almost stochastic dominance}).
Table~\ref{tab:significance-testing} reports the ASO scores 
with a confidence level of $\alpha = 0.05$ adjusted by using the Bonferroni correction~\cite{bonferroni1936teoria}.
See Section~\ref{sec:experiments} for the setup details.

\begin{table}
\centering
\resizebox{0.8\columnwidth}{!}{
\begin{tabular}{r|c|c|c|c|c|c}
\toprule
& \rotatebox[origin=l]{90}{2A [fastText]\textsuperscript{*}} & \rotatebox[origin=l]{90}{2A w/o CR [fastText]\textsuperscript{*}} & \rotatebox[origin=l]{90}{2A w/o CR [BERT]\textsuperscript{*}} & \rotatebox[origin=l]{90}{2A w/o CR [SciBERT]\textsuperscript{*}} & \rotatebox[origin=l]{90}{2A [SciBERT]\textsuperscript{$\dagger$}} & \rotatebox[origin=l]{90}{2A w/o CR [SciBERT]\textsuperscript{$\dagger$}}\\
\midrule
2A [fastText]\textsuperscript{*} & - & 1.0 & \textbf{0.0} & 1.0 & 1.0 & 1.0 \\
\midrule
2A w/o CR [fastText]\textsuperscript{*} & \textbf{0.0} & - & \textbf{0.0} & 1.0 & 1.0 & 1.0 \\
\midrule
2A w/o CR [BERT]\textsuperscript{*} & 1.0 & 1.0 &  - & 1.0 & 1.0 & 1.0 \\
\midrule
2A w/o CR [SciBERT]\textsuperscript{*} & \textbf{0.0} & \textbf{0.0} & \textbf{0.0} & - & 1.0 & 1.0 \\
\midrule
2A [SciBERT]\textsuperscript{$\dagger$} & \textbf{0.0} & \textbf{0.0} & \textbf{0.0} & \textbf{0.0} & -  & 1.0\\
\midrule
2A w/o CR [SciBERT]\textsuperscript{$\dagger$} & \textbf{0.0} & \textbf{0.0} & \textbf{0.0} & \textbf{0.0} & \textbf{0.0} & - \\
\bottomrule
\end{tabular}
}
\caption{ASO scores of the main experimental setups described in Section~\ref{sec:experiments}. (*) CNN model. ($\dagger$) full fine-tuned transformer model. Read as row $\rightarrow$ column.}
\label{tab:significance-testing}
\end{table}

\section{Transformer setups}
\label{sec:appendix-machamp}

The MaChAmp toolkit \cite{van-der-goot-etal-2021-massive} allows for a flexible amount of textual inputs (separated by the \texttt{[SEP]} token) to train the transformer and test the fine-tuned model on. We used SciBERT \cite{beltagy-etal-2019-scibert} and tested the following input configurations:
\begin{enumerate}
    \item The two entities: \\
    \texttt{[} \textit{ent-1} \texttt{[SEP]} \textit{ent-2} \texttt{]}
    
    \item The sentence containing the two entities: \\
    \texttt{[} \textit{sentence} \texttt{]}
    
    \item  The two entities and the sentence containing them: \\
    \texttt{[} \textit{ent-1} \texttt{[SEP]} \textit{ent-2} \texttt{[SEP]} \textit{sentence} \texttt{]}
    
    \item For the third setup, we introduce a marker between the two entities, resulting in a 2-inputs configuration: \\
    \texttt{[} \textit{ent-1} \texttt{[MARK]} \textit{ent-2} \texttt{[SEP]} \textit{sentence} \texttt{]}
    
    \item Finally---following~\citet{baldini-soares-etal-2019-matching}---we augment the input sentence with four word pieces to mark the beginning and the end of each entity mention (\texttt{[E1-START]}, \texttt{[E1-END]}, \texttt{[E2-START]}, \texttt{[E2-END]}): \\
    \texttt{[} \textit{sentence-with-entity-markers} \texttt{]}
\end{enumerate}

Table~\ref{tab:machamp-setups} reports the results of the experiments using MaChAmp on the setups described above.

\begin{table}
\resizebox{\columnwidth}{!}{
\centering
\begin{tabular}{lccccc}
\toprule
$\downarrow$\textbf{Test} | \textbf{Input Setup} $\rightarrow$ & \circled{1} & \circled{2} & \circled{3} &  \circled{4} & \circled{5} \\
\midrule
\semeval{} NLP & 58.15 & 42.08 & 77.79 &	74.85 &	75.12 \\
\scierc{} NLP & 51.42 & 42.16 & 69.90 & 69.09 &71.32  \\
\scierc{} AI-ML & 54.63 & 40.35 & 76.80 & 75.08 & 74.93 \\
\scierc{} CV & 53.16 & 41.09 & 76.11 & 74.73 & 74.21 \\
\scierc{} SPEECH & 49.59 & 40.42 & 67.21 & 66.78 & 67.56 \\
\midrule
\textbf{avg.} & 53.39 & 41.22 & \textbf{73.56} & 72.11 & 72.63 \\
\bottomrule
\end{tabular}
}
\caption{\label{tab:machamp-setups}
Macro F1-scores of the RC using SciBERT \cite{beltagy-etal-2019-scibert} within the MaChAmp toolkit~\cite{van-der-goot-etal-2021-massive}. Setups 1-5 described in Appendix~\ref{sec:appendix-machamp}.
}
\end{table}

\section{Scientific Sub-domain Analysis}
\label{sec:appendix-pca}

\begin{figure}
\begin{subfigure}[t]{\columnwidth}
\centering
\caption{AI-ML (52 abstracts)}
\includegraphics[width=\textwidth]{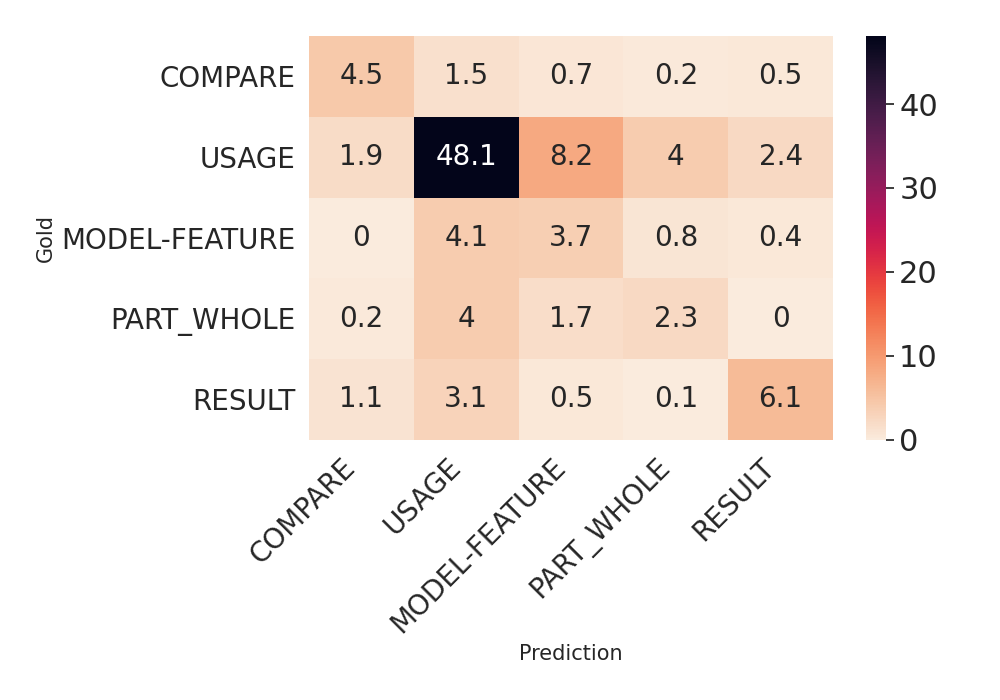}
\label{fig:cm-ai}
\end{subfigure}
\begin{subfigure}[t]{\columnwidth}
\centering
\caption{CV (105 abstracts)}
\includegraphics[width=\textwidth]{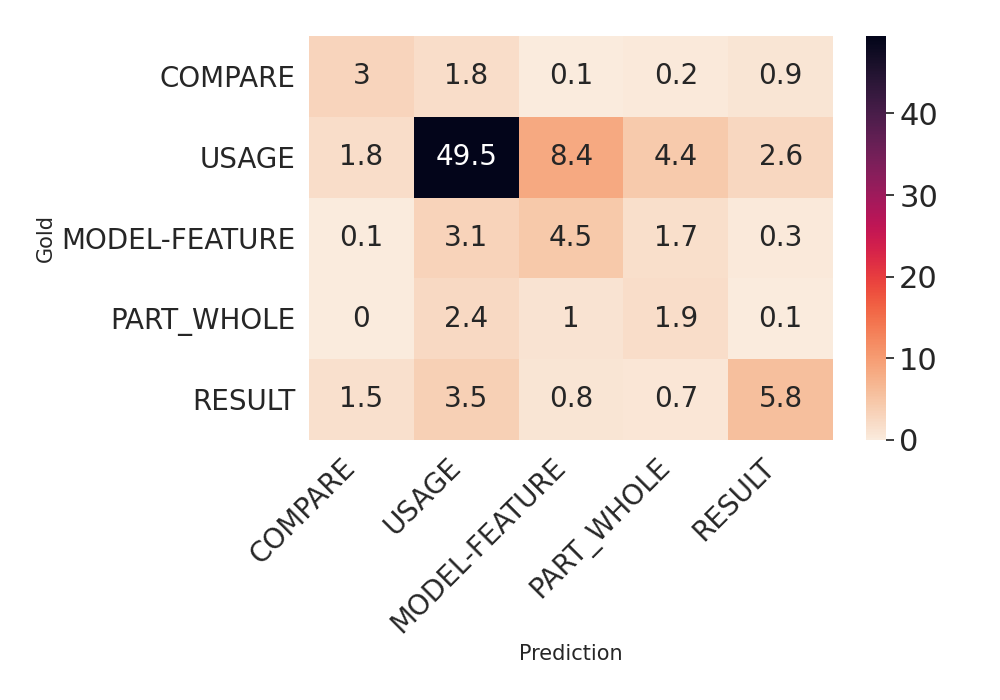}
\label{fig:cm-cv}
\end{subfigure}
\begin{subfigure}[t]{\columnwidth}
\centering
\caption{SPEECH (35 abstracts)}
\includegraphics[width=\textwidth]{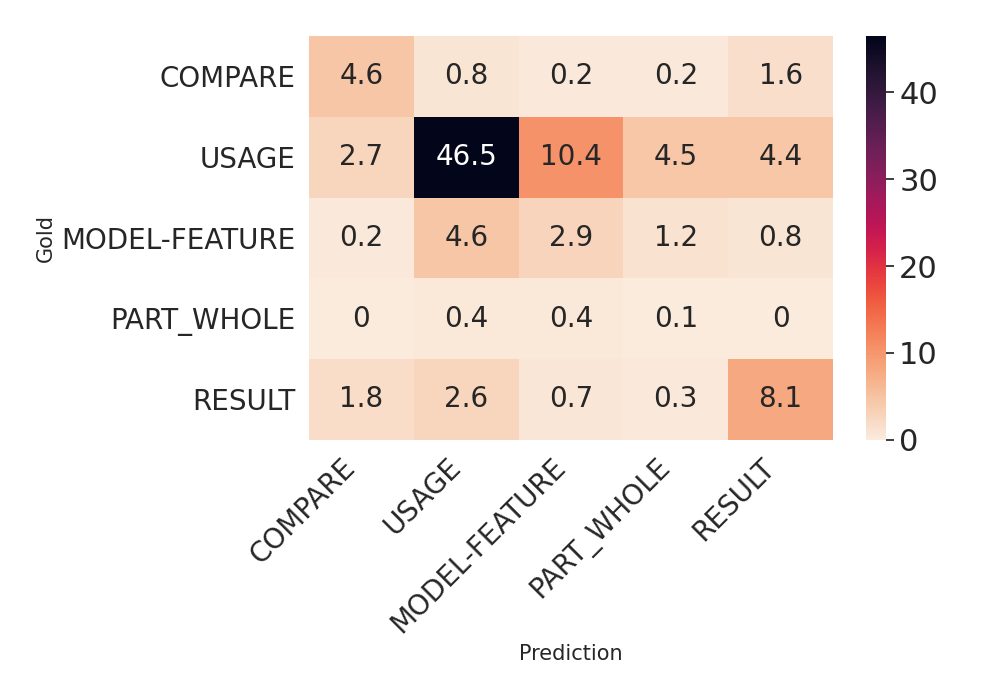}
\label{fig:cm-speech}
\end{subfigure}
\caption{Percentage confusion matrices of the CNN on {\scshape SciERC} sub-domains.}
\label{fig:cm-sub-domains}
\end{figure}

Figure~\ref{fig:cm-sub-domains} contains the confusion matrices of the CNN trained with SciBERT for the AI-ML, CV and SPEECH sub-domains. For fair comparison between the different data amounts the numbers reported are percentages.

\section{Conference Classifier}
\label{sec:appendix-cm}

\begin{figure}[h]
\centering
\includegraphics[width=0.8\columnwidth]{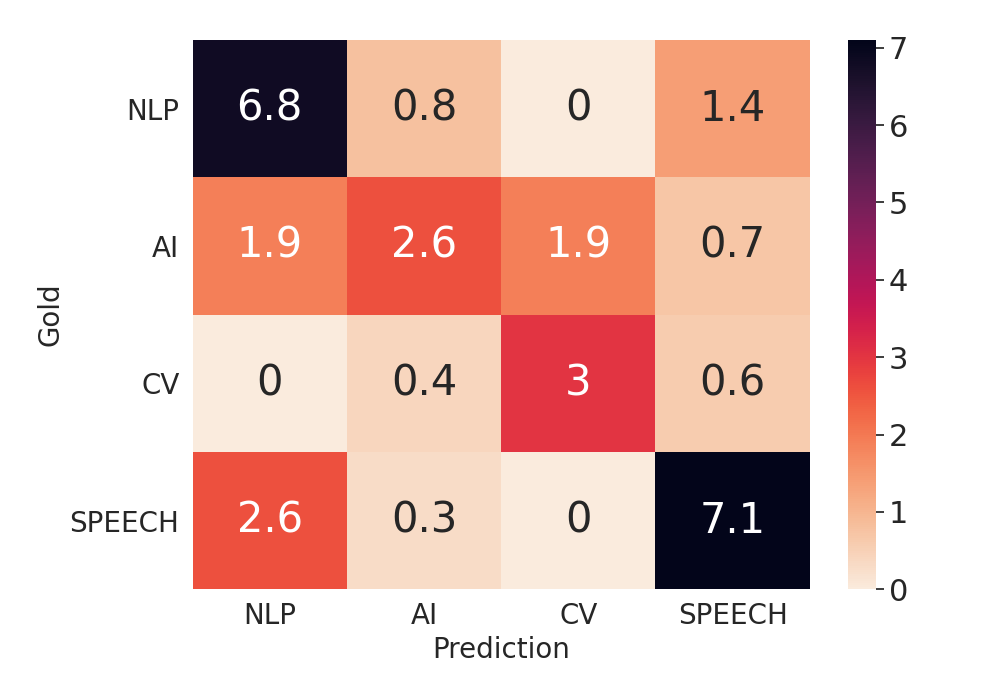}
\caption{Confusion matrix of the conference classification experiment. The numbers reported are the average over three runs on different seeds.}
\label{fig:cm-conf-class}
\end{figure}

Figure~\ref{fig:cm-conf-class} represents the confusion matrix relative to the conference classifier described in Section~\ref{sec:sub-domain-analysis}.

\end{document}